\pdfoutput=1
\PassOptionsToPackage{table,xcdraw,dvipsnames}{xcolor}

\documentclass[11pt]{article}

\usepackage[final]{acl}

\usepackage{times}
\usepackage{latexsym}
\usepackage[edges]{forest}
\usepackage{rotating}

\usepackage[T1]{fontenc}

\usepackage[utf8]{inputenc}

\usepackage{microtype}

\usepackage{inconsolata}

\usepackage{lipsum}
\usepackage{enumitem}
\usepackage{graphicx}
\usepackage{tabularx}  
\usepackage{array}     
\usepackage{booktabs} 
\usepackage{arydshln}  
\usepackage{fancyhdr}
\usepackage{caption}
\usepackage{float}
\usepackage{multicol}
\usepackage{longtable}
\usepackage{geometry}
\usepackage{pdflscape}
\usepackage{ragged2e}
\usepackage{pifont} 
\usepackage{forest}
\usepackage{multirow}
\usepackage{chngcntr}
\usepackage{tikz}
\usepackage{hyperref}
\usepackage{caption}
\usepackage{amsmath}
\usepackage{etoolbox}
\usepackage{tcolorbox}
\usepackage{threeparttable}
\usepackage{titlesec}
\usepackage{titletoc}
\usepackage{xcolor}
\usepackage{hyperref}
\usepackage{changepage}
\usepackage{adjustbox}

\usepackage{minitoc}

\newcommand{\paper}[2]{#1~\citep{#2}}
\newcommand{\codelink}[1]{\href{#1}{Link}}

\makeatletter

\newlength{\appLabelW}
\newlength{\appPnumW}
\newlength{\appSubIndent}
\definecolor{AppIdxRule}{gray}{0.75}

\newcommand{\app@numberline}[1]{%
  \hb@xt@\appLabelW{\bfseries #1\hfil}%
}

\newcommand{\printappendixindex}{%
  \clearpage
  \onecolumn
  \thispagestyle{plain}
  \begingroup
    \@ifpackageloaded{hyperref}{\hypersetup{hidelinks}}{}%

    \def\@dotsep{1.8}

    \setlength{\@pnumwidth}{\appPnumW}
    \setlength{\@tocrmarg}{\appPnumW}

    \def\l@section##1##2{%
      \vspace{7pt}%
      \@dottedtocline{1}{0pt}{\the\appLabelW}{##1}{##2}%
    }
    \def\l@subsection##1##2{%
      \vspace{5pt}%
      \@dottedtocline{2}{\appSubIndent}{\the\appLabelW}{##1}{##2}%
    }

    \setcounter{tocdepth}{2}
    \parindent=0pt
    \parskip=0pt
    \normalsize

    \begin{center}
      {\Large\bfseries Appendix Index}\par
      \vspace{0.35em}
      {\footnotesize\itshape This index lists all appendix sections and their starting pages.}
    \end{center}

    \vspace{0.8em}
    \noindent\textcolor{AppIdxRule}{\rule{\linewidth}{0.45pt}}
    \vspace{0.9em}

    \@starttoc{appx}%
  \endgroup
  \clearpage
  \twocolumn
}

\newcommand{\appsection}[1]{%
  \section{#1}%
  \addcontentsline{appx}{section}{\protect\app@numberline{\thesection}#1}%
}
\newcommand{\appsubsection}[1]{%
  \subsection{#1}%
  \addcontentsline{appx}{subsection}{\protect\app@numberline{\thesubsection}#1}%
}

\makeatother

\title{LLM-Based Human-Agent Collaboration and Interaction Systems: A Survey}

\author{
Henry Peng Zou\textsuperscript{1, *},
Wei-Chieh Huang\textsuperscript{1, *},
Yaozu Wu\textsuperscript{2, *},
Jizhou Guo\textsuperscript{1},
Yankai Chen\textsuperscript{4,5, \dag},  \\ \bf
Chunyu Miao\textsuperscript{1},
Hoang Nguyen\textsuperscript{1}, 
Yue Zhou\textsuperscript{1}, 
\textbf{Weizhi Zhang}\textsuperscript{1},
\textbf{Liancheng Fang}\textsuperscript{1}, \\ \bf
\textbf{Hanrong Zhang}\textsuperscript{1}, 
\textbf{Fangxin Wang}\textsuperscript{1}, 
\textbf{Pengfei Zhang}\textsuperscript{6},
\textbf{Huacan Wang},
\textbf{Langzhou He}\textsuperscript{1},  \\ \bf
\textbf{Yangning Li}\textsuperscript{3},
\textbf{Dongyuan Li}\textsuperscript{2},
\textbf{Renhe Jiang}\textsuperscript{2},
\textbf{Xue Liu}\textsuperscript{4,5}
\textbf{Philip S. Yu}\textsuperscript{1, \dag}
\\
\textsuperscript{1}University of Illinois Chicago,
\textsuperscript{2}University of Tokyo,
\textsuperscript{3}Tsinghua University,\\
\textsuperscript{4}MBZUAI,
\textsuperscript{5}McGill University,
\textsuperscript{6}University of California Irvine\\
\texttt{\{pzou3, whuang80, psyu\}@uic.edu}, \texttt{yaozuwu279@gmail.com}, \texttt{yankaichen@acm.org}
}

\def\domaingaming{Gaming}
\def\domainembodied{Embodied AI}
\def\domainsoftware{Software Development}
\def\domainconversation{Conversational Systems}
\def\domainfinance{Finance}

\begin{document}
\doparttoc 
\faketableofcontents 

\maketitle

\renewcommand{\thefootnote}{}\footnote{$^*$ Equal Contribution. $^\dag$ Corresponding Author.}

\begin{abstract}
Recent advances in large language models (LLMs) have sparked growing interest in building fully autonomous agents. 
However, fully autonomous LLM-based agents still face significant challenges, including limited reliability due to hallucinations, difficulty in handling complex tasks, and substantial safety and ethical risks, all of which limit their feasibility and trustworthiness in real-world applications.
To overcome these limitations, LLM-based human-agent systems (LLM-HAS) incorporate human-provided information, feedback, or control into the agent system to enhance system performance, reliability, and safety. These human-agent collaboration systems enable humans and LLM-based agents to collaborate effectively by leveraging their complementary strengths.
This paper provides the first comprehensive and structured survey of LLM-HAS. It clarifies fundamental concepts, systematically presents core components shaping these systems, including environment and profiling, human feedback, interaction types, orchestration, and communication, explores emerging applications, and discusses unique challenges and opportunities arising from human-AI collaboration.
By consolidating current knowledge and offering a structured overview, we aim to foster further research and innovation in this rapidly evolving interdisciplinary field.
Paper lists and resources are available at \href{https://github.com/HenryPengZou/Awesome-Human-Agent-Collaboration-Interaction-Systems}{Github Repository}. \footnote{\href{https://github.com/HenryPengZou/Awesome-Human-Agent-Collaboration-Interaction-Systems}{https://github.com/HenryPengZou/Awesome-Human-Agent-Collaboration-Interaction-Systems}}
\end{abstract}


\section{Introduction}
Recent advances in Large Language Models (LLMs) have led to growing enthusiasm for building fully autonomous agent systems that use LLMs as a central engine to perceive environments, make decisions, and execute actions to achieve goals~\cite{wang2024survey, li2024survey, zhang2026evoskills}. These agents are often equipped with modules for memory, planning, and tool use, aiming to automate complex workflows with minimal human involvement~\cite{xie2024can, xi2025rise,huang2026rethinking}. However, the pursuit of \emph{full autonomy} faces critical hurdles. \textbf{\textit{(1) Reliability}} remains a major concern due to LLMs' propensity for hallucination, generating plausible but factually incorrect or nonsensical outputs, which undermines trust and can lead to significant errors, especially when actions are chained \cite{gosmar2025hallucination, xu2024reducing, glickman2025human}. \textbf{\textit{(2) Complexity}} often stalls autonomous agents; they struggle with very complicated tasks requiring deep domain expertise, long multi-step execution, nuanced reasoning, dynamic adaptation, or strict long-context consistency dependencies, as seen in scientific research \cite{feng-etal-2024-large, yehudai2025survey}. \textbf{\textit{(3) Safety and Ethical Risks}} escalate with autonomy; agents can take unintended harmful actions, amplify societal biases present in training data, or create accountability gaps, particularly in critical decision-making scenarios involving finance, healthcare, or security \cite{mitchell2025fully, 2024arXiv240605392D, wang2024safe}.

\begin{figure*}[!th]
  \centering
    \includegraphics[width=\textwidth]{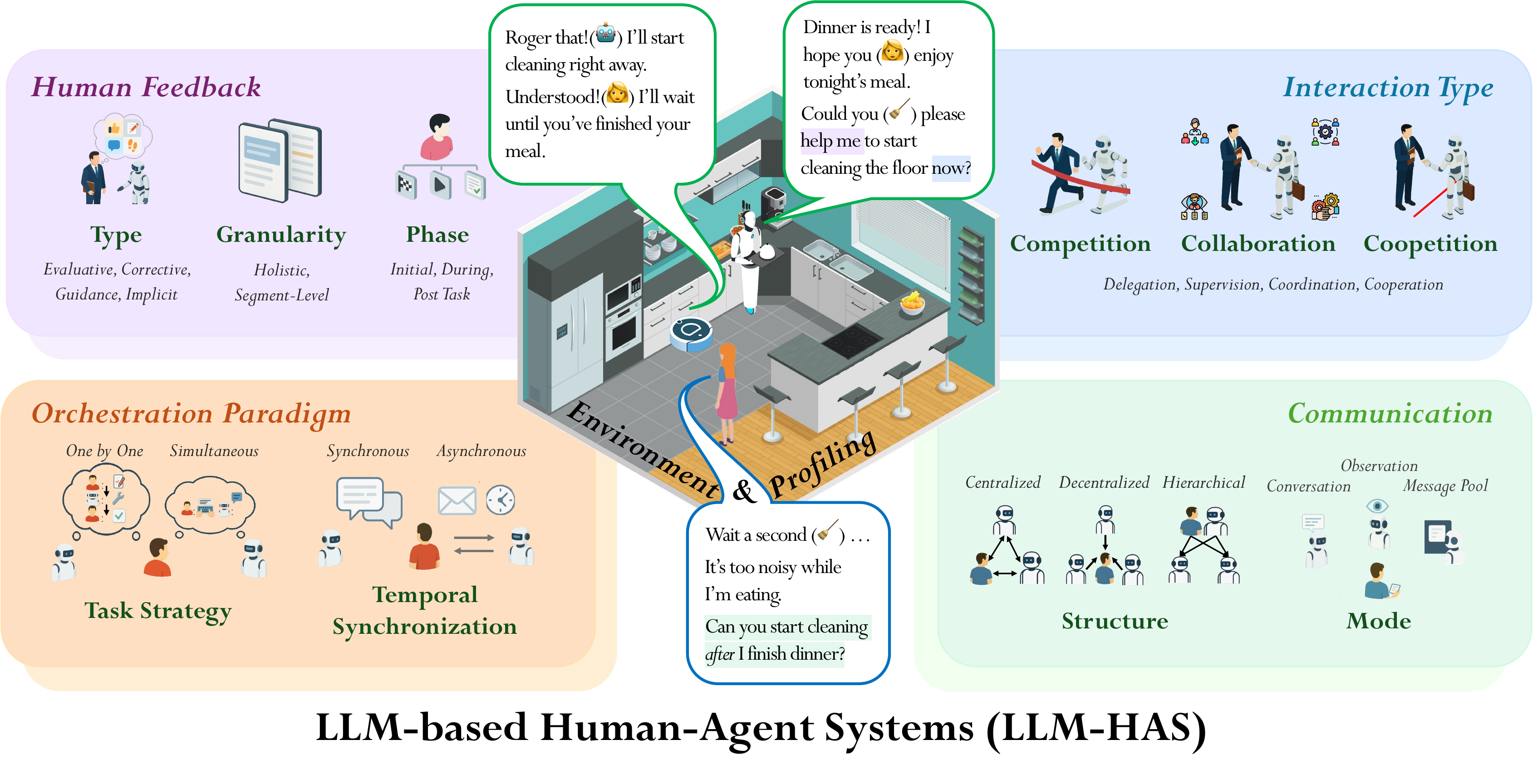}
  \vspace{-9mm}
  \caption{Overview of LLM-based Human-Agent Systems (LLM-HAS). 
  LLM-HAS are interactive frameworks where humans actively provide additional information, feedback, or control during interaction with an LLM-powered agent to enhance system performance, reliability, and safety.
  The system is composed of five core components: 
  \textcolor{Black}{\textbf{Environment \& Profiling}} (including environment settings, and role definitions, goals, and agent capabilities such as planning and memory), \textcolor{Mulberry}{\textbf{Human Feedback}} (with varying types, timing, and granularity), \textcolor{RoyalBlue}{\textbf{Interaction Types}} (collaborative, competitive, cooperative, or mixed), \textcolor{BurntOrange}{\textbf{Orchestration}} (task strategy and temporal synchronization), and \textcolor{ForestGreen}{\textbf{Communication}} (information flow structure and mode). 
  }
  \label{fig:overview_llm_has}
\end{figure*}


The persistence of these challenges suggests that full autonomy may be unsuitable for many real-world applications \cite{mitchell2025fully, natarajan2025human} and underscores a crucial insight often overlooked in the drive for pure automation: the indispensable role of human involvement. Humans are frequently needed to provide additional information, essential clarification, or domain knowledge, offer vital feedback and corrections, and exercise necessary oversight and control. These motivate a paradigm shift towards systems explicitly designed for human-agent collaboration: \textbf{\textit{LLM-based Human-Agent Systems} (LLM-HAS)}.

While surveys on LLM-based autonomous agents \cite{wang2024survey, li2024survey}, multi-agent systems \cite{tran2025multi, wu2025multi}, and specific applications exist \cite{wang2025survey, peng2025survey}, a dedicated synthesis focusing specifically on LLM-based human-agent systems is lacking. This survey fills the gap by providing a comprehensive and structured overview of the LLM-HAS. It clarifies the fundamental concepts (Section \ref{sec:llm_has_def}) and systematically presents its core components (Section \ref{sec:components}), major implementation strategies (Section \ref{sec:implement_strategies}), emerging applications (Section \ref{sec:application_resource}), open-source frameworks, datasets and benchmarks (Section \ref{sec:implementation}), and unique challenges and opportunities (Section \ref{sec:challenge}) within this specific niche. To the best of our knowledge, this is still the first survey on LLM-based human-agent systems. 
We aim to consolidate current knowledge and inspire further research and application in this rapidly evolving field. We maintain an open-source \href{https://github.com/HenryPengZou/Awesome-Human-Agent-Collaboration-Interaction-Systems}{GitHub repository} to provide a sustainable resource complementing our survey paper.


\definecolor{hidden-draw}{RGB}{0,0,0}
\definecolor{HumanFeedbackClr}{RGB}{230,210,255}
\colorlet   {HumanFeedbackLight}{HumanFeedbackClr!55}

\definecolor{interactClr}{RGB}{190,215,255}
\colorlet   {interactLight}{interactClr!55}

\definecolor{orchestClr}{RGB}{255,196,141}
\colorlet   {orchestLight}{orchestClr!55}

\definecolor{communClr}{RGB}{206,240,220}
\colorlet   {communLight}{communClr!55}

\tikzstyle{my-box}=[
    rectangle,
    draw=hidden-draw,
    rounded corners,
    text opacity=1,
    minimum height=1.5em,
    minimum width=5em,
    inner sep=2pt,
    align=center,
    fill opacity=.5,
    line width=0.8pt,
]
\tikzstyle{leaf}=[my-box, minimum height=1.5em,
    fill=white, text=black, align=left,font=\tiny,
    inner xsep=2pt,
    inner ysep=4pt,
    line width=0.8pt,
]
\begin{figure*}[!t]
    \centering
    \resizebox{0.9\textwidth}{!}{
        \begin{forest}
            forked edges,
            for tree={
                grow=east,
                reversed=true,
                anchor=base west,
                parent anchor=east,
                child anchor=west,
                base=left,
                font=\footnotesize,
                anchor=center,
                align=center, 
                text centered,
                rectangle,
                draw=hidden-draw,
                rounded corners,
                align=left,
                minimum width=2em,
                edge+={darkgray, line width=1pt},
                s sep=3pt,
                inner xsep=2pt,
                inner ysep=3pt,
                line width=0.8pt,
                ver/.style={rotate=90, child anchor=north, parent anchor=south, anchor=center, font=\LARGE\bfseries},
            }
            [
                LLM-based Human-Agent Systems, ver
                [
                    Human \\ Feedback \\ (Sec~\ref{Sec:Human Feedback}), fill= HumanFeedbackClr, text width= 7em, anchor=center, align=center, font=\large
                    [
                        Feedback \\ Type, fill=HumanFeedbackLight, text width= 5.5em, anchor=center, align=center, font=\large
                        [
                            Evaluative, fill=HumanFeedbackLight, text width= 5.5em, anchor=center, align=center, font=\large
                            [    
                                AgentCoord~\cite{pan2024agentcoord}{,} CowPilot~\cite{huq2025cowpilotframeworkautonomoushumanagent}{,}\\ ConvCodeWorld~\cite{han2025convcodeworld}{,} GDfC~\cite{wang2025human}{,} DigiRL~\cite{NEURIPS2024_1704ddd0}, leaf, fill=HumanFeedbackLight, text width=45em, anchor=center, align=center, text centered, font = \normalsize
                            ]
                        ]
                        [
                            Corrective, fill=HumanFeedbackLight, text width= 5.5em, anchor=center, align=center, font=\large
                            [    
                                Collaborative Gym~\cite{shao2024collaborative}{,} ReHAC~\cite{feng-etal-2024-large}{,}\\AI Chains~\cite{10.1145/3491102.3517582}{,} HRC DMP~\cite{liu2024enhancing}{,} BPMN~\cite{ait2024towards}, leaf, fill=HumanFeedbackLight, text width=45em, anchor=center, align=center, text centered, font = \normalsize
                            ]
                        ]
                        [
                            Guidance, fill=HumanFeedbackLight, text width= 5.5em, anchor=center, align=center, font=\large
                            [    
                                FineArena~\cite{xu2025finarena}{,} InteractGen~\cite{suninteractgen}{,}\\ConvCodeWorld~\cite{han2025convcodeworld}{,} DPT Agent~\cite{zhang2025leveraging}{,} Organized Teams~\cite{guo2024embodiedllmagentslearn}, leaf, fill=HumanFeedbackLight, text width=45em, anchor=center, align=center, text centered, font = \normalsize
                            ]
                        ]
                        [
                            Implicit, fill=HumanFeedbackLight, text width= 5.5em, anchor=center, align=center, font=\large
                            [    
                                Prison Dilemm~\cite{jiang2025experimentalexplorationinvestigatingcooperative}{,} AXIS~\cite{lu2024turn}{,} AssistantX~\cite{sun2024assistantx}{,}\\EmoAgent~\cite{qiu2025emoagent}, leaf, fill=HumanFeedbackLight, text width=45em, anchor=center, align=center, text centered, font = \normalsize
                            ]
                        ]
                    ]
                    [
                        Feedback \\ Granularity, fill= HumanFeedbackLight, text width= 5.5em, anchor=center, align=center, font=\large
                        [
                            Holistic, fill=HumanFeedbackLight, text width= 5.5em, anchor=center, align=center, font=\large
                            [    
                                Help Feedback~\cite{mehta-etal-2024-improving}{,} PARTNR~\cite{chang2024partnr}{,} A2C~\cite{tariq2025a2c}{,}\\CoELA~\cite{zhang2024building}{,} DPT Agent~\cite{zhang2025leveraging}, leaf, fill=HumanFeedbackLight, text width=45em, anchor=center, align=center, text centered, font =\normalsize
                            ]
                        ]
                        [
                            Segment, fill=HumanFeedbackLight, text width= 5.5em, anchor=center, align=center, font=\large
                            [    
                                CowPilot~\cite{huq2025cowpilotframeworkautonomoushumanagent}{,} SWEET-RL~\cite{zhou2025sweetrltrainingmultiturnllm}{,} PDFChatAnnotator~\cite{10.1145/3640543.3645174}{,} \\ TaPA~\cite{wu2023embodied}, leaf, fill=HumanFeedbackLight, text width=45em, anchor=center, align=center, text centered, font = \normalsize
                            ]
                        ]
                    ]
                    [
                        Feedback \\ Phase, fill= HumanFeedbackLight, text width= 5.5em, anchor=center, align=center, font=\large
                        [
                            Initial Task, fill=HumanFeedbackLight, text width= 5.5em, anchor=center, align=center, font=\large
                            [    
                                AgentCoord~\cite{pan2024agentcoord}{,} AssistantX~\cite{sun2024assistantx}{,} Hierarchical Agent\cite{Liu2023LLMPoweredHL}, leaf, fill=HumanFeedbackLight, text width=45em, anchor=center, align=center, text centered, font = \normalsize
                            ]
                        ]
                        [
                            During Task, fill=HumanFeedbackLight, text width= 5.5em, anchor=center, align=center, font=\large
                            [    
                                ConvCodeWorld~\cite{han2025convcodeworld}{,} HRC Manipulation~\cite{liu2023llm}{,} A2C~\cite{tariq2025a2c}, leaf, fill=HumanFeedbackLight, text width=45em, anchor=center, align=center, text centered, font = \normalsize
                            ]
                        ]
                        [
                            Post-Task, fill=HumanFeedbackLight, text width= 5.5em, anchor=center, align=center, font=\large
                            [    
                               GDfC~\cite{wang2025human}{,} SOTOPIA~\cite{zhou2024sotopia}{,}DigiRL~\cite{NEURIPS2024_1704ddd0}{,}\\ EmoAgent~\cite{qiu2025emoagent}, leaf, fill=HumanFeedbackLight, text width=45em, anchor=center, align=center, text centered, font = \normalsize
                            ]
                        ]
                    ]
                ]
                [
                    Interaction \\ Type \\ (Sec~\ref{sec:interaction types}), fill=interactClr, text width= 7em, anchor=center, align=center, font=\large
                    [
                        Collaboration, fill=interactLight, text width= 6em, anchor=center, align=center, font=\large
                        [
                            Supervision, fill=interactLight, text width= 6em, anchor=center, align=center, font=\normalsize
                            [    
                                EasyLAN~\cite{pan2024human}{,} CowPilot~\cite{huq2025cowpilotframeworkautonomoushumanagent}{,} Hierarchical Agent~\cite{Liu2023LLMPoweredHL}{,}\\ConvCodeWorld~\cite{han2025convcodeworld}{,} HRC Manipulation~\cite{liu2023llm}, leaf, fill=interactLight, text width=44em, anchor=center, align=center, text centered, font = \normalsize
                            ]
                        ]
                        [
                            Delegation, fill=interactLight, text width= 6em, anchor=center, align=center, font=\large
                            [    
                                Collaborative Gym~\cite{shao2024collaborative}{,} GDfC~\cite{wang2025human}{,} HRC Manufa~\cite{10711843}{,}\\PaLM-E~\cite{10.5555/3618408.3618748}{,} DigiRL~\cite{NEURIPS2024_1704ddd0}, leaf, fill=interactLight, text width=44em, anchor=center, align=center, text centered, font = \normalsize
                            ]
                        ]
                        [
                            Coordination, fill=interactLight, text width= 6em, anchor=center, align=center, font=\large
                            [    
                                MTOM~\cite{Zhang2024MutualTO}{,} Ask Before Plan~\cite{zhang-etal-2024-ask}{,} MindAgent ~\cite{gong2023mindagent}{,}\\MetaGPT ~\cite{hong2023metagpt}, leaf, fill=interactLight, text width=44em, anchor=center, align=center, text centered, font = \normalsize
                            ]
                        ]
                        [
                            Cooperation, fill=interactLight, text width= 6em, anchor=center, align=center, font=\large
                            [    
                                CoELA~\cite{zhang2024building}{,} InteractGen~\cite{suninteractgen}{,} PARTNR~\cite{chang2024partnr}{,} \\MAIH~\cite{wang2024safe}, leaf, fill=interactLight, text width=44em, anchor=center, align=center, text centered, font = \normalsize
                            ]
                        ]
                    ]
                    [
                        Competition, fill=interactLight, text width= 6em, anchor=center, align=center, font=\large
                        [    
                            SOTOPIA~\cite{zhou2024sotopia}, leaf, fill=interactLight, text width=51.6em, anchor=center, align=center, text centered, font = \normalsize
                        ]
                    ]
                    [
                        Coopetition, fill=interactLight, text width= 6em, anchor=center, align=center, font=\large
                        [    
                            Prison Dilemm~\cite{jiang2025experimentalexplorationinvestigatingcooperative}{,} SOTOPIA~\cite{zhou2024sotopia}, leaf, fill=interactLight, text width=51.6em, anchor=center, align=center, text centered, font = \normalsize
                        ]
                    ]
                ]
                [
                    Orchestration \\ Paradigm \\ (Sec~\ref{sec:orchestration paradigm}), fill=orchestClr, text width= 7em, anchor=center, align=center, font=\large
                    [
                        Task \\ Strategy, fill=orchestLight, text width= 7.2em, anchor=center, align=center, font=\large
                        [
                            One by One, fill=orchestLight, text width= 6.5em, anchor=center, align=center, font=\large
                            [    
                                WebLINX~\cite{lu2024weblinx}{,} Co-STORM~\cite{jiang-etal-2024-unknown}{,} Agency Task~\cite{sharma2024investigatingagencyllmshumanai}{,}\\Help Feedback~\cite{mehta-etal-2024-improving}{,} Ask Before Plan~\cite{zhang-etal-2024-ask}, leaf, fill=orchestLight, text width=42.2em, anchor=center, align=center, text centered, font = \normalsize
                            ]
                        ]
                        [
                            Simultaneous, fill=orchestLight, text width= 6.5em, anchor=center, align=center, font=\large
                            [    
                                MTOM~\cite{Zhang2024MutualTO}{,} HRT-ML~\cite{liu2024effect}{,} DPT Agent~\cite{zhang2025leveraging}{,}\\CoELA~\cite{zhang2024building}, leaf, fill=orchestLight, text width=42.2em, anchor=center, align=center, text centered, font = \normalsize
                            ]
                        ]
                    ]
                    [
                        Temporal \\ Synchronization, fill=orchestLight, text width= 7.2em, anchor=center, align=center, font=\large
                        [
                            Synchronous, fill=orchestLight, text width= 6.5em, anchor=center, align=center, font=\large
                            [    
                                AgentCoord~\cite{pan2024agentcoord}{,} HRC Assembly~\cite{gkournelos2024llm}{,}\\HRC Manufa~\cite{10711843}{,} UserBench~\cite{Qian2025UserBenchAI}{,} MUA-RL~\cite{Zhao2025MUARLMU}, leaf, fill=orchestLight, text width=42.2em, anchor=center, align=center, text centered, font = \normalsize
                            ]
                        ]
                        [
                            Asynchronous, fill=orchestLight, text width= 6.5em, anchor=center, align=center, font=\large
                            [    
                                Collaborative Gym ~\cite{shao2024collaborative}{,} MetaGPT~\cite{hong2023metagpt}{,} TaPA~\cite{wu2023embodied}, leaf, fill=orchestLight, text width=42.2em, anchor=center, align=center, text centered, font = \normalsize
                            ]
                        ]
                    ]   
                ]
                [
                    Communication \\ (Sec~\ref{sec:communication}), fill=communClr, text width= 7em, anchor=center, align=center, font=\large
                    [
                        Structure, fill=communLight, text width= 5em, anchor=center, align=center, font=\large
                        [
                            Centralized, fill=communLight, text width= 6em, anchor=center, align=center, font=\large
                            [    
                                Drive As You Speack~\cite{cui2024drive}{,} Organized Teams~\cite{guo2024embodiedllmagentslearn}{,} \\ HRC Manufa~\cite{10711843}{,} DigiRL~\cite{NEURIPS2024_1704ddd0}, leaf, fill=communLight, text width=45em, anchor=center, align=center, text centered, font = \normalsize
                            ]
                        ]
                        [
                            Decentralized, fill=communLight, text width= 6em, anchor=center, align=center, font=\large
                            [    
                                Prison Dilemm~\cite{jiang2025experimentalexplorationinvestigatingcooperative}{,} InteractGen~\cite{suninteractgen}{,} TaPA~\cite{wu2023embodied}{,}\\MINT~\cite{wang2024mint}, leaf, fill=communLight, text width=45em, anchor=center, align=center, text centered, font = \normalsize
                            ]
                        ]
                        [
                            Hierarchical, fill=communLight, text width= 6em, anchor=center, align=center, font=\large
                            [    
                                FineArena~\cite{xu2025finarena}{,} AgentCoord ~\cite{pan2024agentcoord}{,}\\EasyLAN~\cite{pan2024human}{,} Hierarchical Agent~\cite{Liu2023LLMPoweredHL}{,} EmoAgent~\cite{qiu2025emoagent}, leaf, fill=communLight, text width=45em, anchor=center, align=center, text centered, font = \normalsize
                            ]
                        ]
                    ]
                    [
                        Mode, fill=communLight, text width= 5em, anchor=center, align=center, font=\large
                        [
                            Conversation, fill=communLight, text width= 6em, anchor=center, align=center, font=\large
                            [    
                                DigiRL~\cite{NEURIPS2024_1704ddd0}{,} WebLINX~\cite{lu2024weblinx}{,} Organized Teams~\cite{guo2024embodiedllmagentslearn}{,}\\ConvCodeWorld~\cite{han2025convcodeworld}{,} Help Feedback~\cite{mehta-etal-2024-improving}, leaf, fill=communLight, text width=45em, anchor=center, align=center, text centered, font = \normalsize
                            ]
                        ]
                        [
                            Observation, fill=communLight, text width= 6em, anchor=center, align=center, font=\large
                            [    
                                AXIS~\cite{lu2024turn}{,} Attentive Supp~\cite{tanneberg2024help}{,} EasyLAN~\cite{pan2024human}{,}\\MineWorld~\cite{guo2025mineworld}, leaf, fill=communLight, text width=45em, anchor=center, align=center, text centered, font = \normalsize
                            ]
                        ]
                        [
                            Message \\ Pool, fill=communLight, text width= 6em, anchor=center, align=center, font=\large
                            [    
                                SMALL~\cite{wang2024safe}{,} HRT-ML~\cite{liu2024effect}{,} MetaGPT~\cite{hong2023metagpt}, leaf, fill=communLight, text width=45em, anchor=center, align=center, text centered, font = \normalsize
                            ]
                        ]
                    ]
                ]
            ]
        \end{forest}
    }
    \caption{Taxonomy of LLM-based Human-Agent Systems. A more detailed and structured categorization of representative works is provided in the appendix (Table \ref{tab:humanfeedback} and \ref{tab:humancommunication}).}
    \label{taxo_of_llmhas}
\end{figure*}

\section{LLM-Based Human-Agent Systems}
\label{sec:llm_has_def}

We define LLM-based human-agent systems as interactive frameworks where humans actively provide additional information, feedback, or control during interaction with an LLM-powered agent to enhance system performance, reliability, and safety \cite{feng-etal-2024-large, shao2024collaborative, mehta-etal-2024-improving}. The core idea is \textbf{synergy}: combining unique human strengths—like intuition, creativity, expertise, ethical judgment, and adaptability—with LLM agent capabilities such as vast knowledge recall, computational speed, and sophisticated language processing. 
LLM-HAS builds upon core LLM agent components but places critical emphasis on the human's interactive role and capabilities:
\begin{enumerate}[label=\textbf{\textit{(\arabic*)}}, parsep=0pt] 
  \item \textbf{\textit{Providing Information / Clarification:}}  
    Humans provide additional information that agents lack or cannot reliably infer, such as login credentials, payment details, domain expertise, constraints, or resolve ambiguities, helping agents interpret situations more accurately \cite{naik2025empirical, kim2025beyond}.
  \item \textbf{\textit{Providing Feedback / Error Correction:}}  
    Humans evaluate agent outputs and provide feedback, ranging from simple ratings to complex critiques, demonstrations or corrections, effectively guiding agents' adjustment \cite{gao2024taxonomy, dutta2024problem,zou2026users}.
  \item \textbf{\textit{Taking Control / Action:}}  
    In high-stakes or sensitive scenarios (e.g., healthcare, privacy, or ethics), humans retain the authority to override, redirect, or halt agent actions, ensuring accountability, safety, and alignment with human values \cite{chen2025reinforcing, natarajan2025human, xiao2023llm}.
\end{enumerate}

Figure \ref{fig:overview_llm_has} provides a generalized overview of LLM‑based human‑agent systems. These systems operate within a defined \textcolor{Black}{\textbf{Environment}} (e.g., physical world, simulation) that provides context and stimuli. \textcolor{Black}{\textbf{Human \& Agent Profiling}} characterizes the participants’ roles and goals, and the agent’s core LLM engine augmented with capabilities like planning, memory, and tool use. \textcolor{Mulberry}{\textbf{Human Feedback}} can occur during different phases in various types and granularities. Human‑Agent \textcolor{RoyalBlue}{\textbf{Interaction Types}} may be collaborative (most common), competitive, cooperative, or mixed. The \textcolor{BurntOrange}{\textbf{Orchestration}} layer governs high‑level coordination, choosing a task strategy (e.g., sequential one‑by‑one versus parallel simultaneous execution) and a temporal synchronization mode (real‑time synchronous exchanges versus delayed asynchronous workflows) so that each actor acts at the right moment. The \textcolor{ForestGreen}{\textbf{Communication}} layer specifies how information flows, defining message structure (centralized, decentralized, hierarchical) and mode (conversation, observation signals, or shared message pools). 
The effective interplay and configuration of these components, along with various human feedback, are critical for tailoring the system to specific tasks and optimizing the overall system's performance.
The taxonomy of LLM-based human-agent systems is outlined in Figure \ref{taxo_of_llmhas}. A detailed and structured categorization of representative works is provided in the Table \ref{tab:humanfeedback} and Table \ref{tab:humancommunication}.

\section{Core Components}
\label{sec:components}
In this section, we examine LLM-HAS through five core aspects: environment \& profiling, human feedback, interaction type, orchestration paradigm, and communication. These dimensions provide a unified standard for analyzing existing work and guiding the design of future systems. 

\subsection{Environment and Profiling}
\label{sec:environment Settings}

\noindent \textbf{Environment Setting.} 
The environment in LLM‑HAS defines a shared interaction space that can exist either in the physical world, such as offices \cite{suninteractgen}, or in fully simulated virtual environments where agents and humans engage under controlled conditions \cite{suninteractgen, zhang2024building, guo2024embodiedllmagentslearn}. These systems can be configured in various ways, including single-human single-agent, single-human multi-agent, multi-human single-agent, and multi-human multi-agent setups, each reflecting different collaboration dynamics and complexities.\\

\noindent \textbf{Human \& Agent Profiling. } Human participants can be broadly categorized as \textit{lazy} or \textit{informative} users. Lazy users provide minimal guidance, typically offering evaluative feedback such as binary correctness or scalar rating. In contrast, informative users engage deeply by offering demonstrations, detailed guidance, refinements, or even taking over parts of the task \cite{wang2024mint, liu2024effect, han2025convcodeworld}. On the other side, agents are profiled by their roles and capabilities, which range from versatile general assistants to specialized experts in mathematics, engineering, medicine, or robotic cleaning, each adapted to the particular demands of its operational context \cite{guo2024large, samuel2024personagym}.

\subsection{Human Feedback}
\label{Sec:Human Feedback}

\definecolor{headercolor}{RGB}{52, 100, 123}          
\definecolor{coralpeach}{RGB}{255, 229, 221}          
\definecolor{coralpeachlight}{RGB}{255, 243, 239}     

\definecolor{gentlebeige}{RGB}{250, 243, 223}         
\definecolor{gentlebeigelight}{RGB}{253, 249, 236}    

\definecolor{coolblue}{RGB}{232, 243, 255}            
\definecolor{coolbluelight}{RGB}{243, 248, 255}       

\definecolor{mistyrose}{RGB}{255, 230, 234}          
\definecolor{mistyroselight}{RGB}{255, 244, 246}     

\definecolor{sagegreen}{RGB}{229, 245, 234}          
\definecolor{sagegreenlight}{RGB}{241, 252, 245}     

\definecolor{lavenderfog}{RGB}{232, 232, 250}        
\definecolor{lavenderfoglight}{RGB}{244, 244, 255}   

\definecolor{softfern}{RGB}{235, 248, 236}
\definecolor{softfernlight}{RGB}{245, 252, 245}

\definecolor{palesand}{RGB}{250, 244, 230}
\definecolor{palesandlight}{RGB}{254, 250, 242}

\definecolor{powdersky}{RGB}{229, 241, 252}
\definecolor{powderskylight}{RGB}{241, 247, 255}

\definecolor{highlightcolor}{RGB}{52, 100, 123}      
\definecolor{examplecolor}{RGB}{120, 160, 180}       

\definecolor{elegantorangelight}{RGB}{255, 235, 215}  
\definecolor{elegantorange}{RGB}{255, 200, 150}  

\definecolor{elegantpurplelight}{RGB}{235, 230, 250}  
\definecolor{elegantpurple}{RGB}{180, 160, 220}  

\definecolor{elegantbrown}{RGB}{100, 75, 50}
\definecolor{elegantdarkpurple}{RGB}{92, 64, 133}  

\definecolor{elegantdarkgreen}{RGB}{44, 89, 66}  
\definecolor{elegantgreen}{RGB}{67, 140, 100}  

\definecolor{communClr}{RGB}{230,210,255}

\renewcommand{\arraystretch}{1.3}

\begin{table*}[!tb]
\centering
\fontsize{6.8 pt}{10pt}\selectfont
\begin{tabularx}{\textwidth}{%
  >{\raggedright\arraybackslash}p{1.5 cm}
  >{\raggedright\arraybackslash}p{1.8 cm}
  >{\raggedright\arraybackslash}p{3.6 cm}
  >{\raggedright\arraybackslash}p{3.6 cm}
  >{\raggedright\arraybackslash}X}
\toprule
\rowcolor{communClr}
\color{black}\textbf{Dimension} & \color{black}\textbf{Category} & \color{black}\textbf{Definition Summary} & \color{black}\textbf{Key Characteristics / Trade-offs} & \color{black}\textbf{Example Work} \\
\midrule

\rowcolor{coralpeachlight}
\textbf{Type} & \textit{Evaluative} & 
User provides an \textbf{assessment} of the agent’s output quality, typically as  \textbf{\textcolor{elegantgreen}{binary assessment}},  \textbf{\textcolor{elegantgreen}{scalar rating}}, or \textbf{\textcolor{elegantgreen}{preference ranking}}. 
& \ding{172}~Easy to collect, scalable.
\ding{173}~Less specific signal for improvement. 
& \textit{EmoAgent}
\citep{qiu2025emoagent}, \textit{MINT} \citep{wang2024mint}, \textit{SOTOPIA} \citep{zhou2024sotopia} \\

\rowcolor{coralpeachlight}
& \textit{Corrective} & User \textbf{offers edits or fixes} to the agent’s behavior. & \ding{172}~Highly informative, clear signal for improvement.
\ding{173}~Higher user effort, often fine-grained \& interactive. & \textit{SymbioticRAG} \citep{sun2025symbioticrag}, \textit{SWEET-RL} \citep{zhou2025sweetrltrainingmultiturnllm}, \textit{AI Chains} \citep{10.1145/3491102.3517582} \\

\rowcolor{coralpeachlight}
& \textit{Guidance} & User proactively provides \textbf{instructions}, \textbf{demonstrations}, or \textbf{critiques} to shape the agent’s behavior. & \ding{172}~Bootstraps learning, conveys complex goals, proactive alignment.
\ding{173}~Requires clear specification from user. & \textit{Drive As You Speack} \citep{cui2024drive}, \textit{Hierarchical Agent}\citep{Liu2023LLMPoweredHL}, \textit{Ask Before Plan} \citep{zhang-etal-2024-ask} \\

\rowcolor{coralpeachlight}
& \textit{Implicit} & \textbf{Inferred by the agent observing user actions or control signals}, rather than explicitly stated or direct output modifications. & 
\ding{172}~Natural, unobtrusive collection. 
\ding{173}~Ambiguous, requires careful interpretation. & \textit{MTOM} \citep{Zhang2024MutualTO}, \textit{Attentive Supp.} \citep{tanneberg2024help}, \textit{MineWorld} \citep{guo2025mineworld} \\

\rowcolor{palesandlight}
\textbf{Granularity} & \textit{Coarse-grained / Holistic} & Single assessment/signal for \textbf{an entire agent} \textbf{\textcolor{elegantgreen}{output}}, \textbf{\textcolor{elegantgreen}{trajectory}}, or \textbf{\textcolor{elegantgreen}{task outcome}}. & 
\ding{172}~ Simple for user, good for overall assessment 
\ding{173}~ Obscures specific errors, less precise learning signal. & \textit{AssistantX} \citep{sun2024assistantx}, \textit{Help Feedback} \citep{mehta-etal-2024-improving}, \textit{AXIS} \citep{lu2024turn} \\

\rowcolor{palesandlight}
& \textit{Fine-grained / Segment-Level} & Feedback targeting \textbf{specific parts of agent} \textbf{\textcolor{elegantgreen}{output}}, \textbf{\textcolor{elegantgreen}{actions}}, or \textbf{\textcolor{elegantgreen}{process}}. & 
\ding{172}~ Precise learning signal, crucial for debugging complex skills
\ding{173}~ Potentially higher user effort/burden. & \textit{Collaborative Gym} \citep{shao2024collaborative}, \textit{Prison Dilemm} \citep{jiang2025experimentalexplorationinvestigatingcooperative}, \textit{FineArena} \citep{xu2025finarena} \\

\rowcolor{powderskylight}
\textbf{Phase} & \textit{Initial Setup \& Goal Definition} & Feedback provided \textbf{before} task execution, \textbf{configuring} the agent system and \textbf{defining} the \textbf{\textcolor{elegantgreen}{task}}, \textbf{\textcolor{elegantgreen}{goals}}, \textbf{\textcolor{elegantgreen}{constraints}}, and \textbf{\textcolor{elegantgreen}{preference}}. & 
\ding{172}~ Initial and proactive alignment, prevents costly errors, sets constraints
\ding{173}~ Requires upfront user input. & \textit{AgentCoord} \citep{pan2024agentcoord}, \textit{GDfC} \citep{wang2025human}, \textit{SMALL} \citep{wang2024safe} \\

\rowcolor{powderskylight}
& \textit{During Task Execution} & Online, interactive feedback \textbf{while the agent is actively performing the task}, enabling \textbf{\textcolor{elegantgreen}{real-time adaptation}}. & 
\ding{172}~ Enables real-time adaptation, crucial for dynamic/collaborative tasks
\ding{173}~ Requires timely notification and responsive interfaces. & \textit{InteractGen} \citep{suninteractgen}, \textit{CowPilot} \citep{huq2025cowpilotframeworkautonomoushumanagent}, \textit{EasyLAN} \citep{pan2024human} \\

\rowcolor{powderskylight}
& \textit{Post-Task Eval. \& Refinement} & Feedback provided \textbf{after task completion} to assess outcomes and \textbf{provide suggestions} for \textbf{\textcolor{elegantgreen}{future use}}. & 
\ding{172}~ Non-disruptive, good for aggregate data/offline learning
\ding{173}~ No impact on completed task. & \textit{HRT-ML} \citep{liu2024effect}, \textit{M3HF} \citep{wang2025m3hf}, \textit{MAIH} \citep{wang2024safe} \\

\bottomrule
\end{tabularx}
\caption{Dimensions of Human Feedback in LLM-based human–agent systems, including feedback type, granularity, and phase. For each dimension, a summary, key characteristics, and example works are provided for comparison. A detailed overview of human feedback types and their subtypes is provided in our appendix (Table~\ref{tab:human_feedback_type}).}
\label{tab:feedback_dimensions}
\end{table*}

\noindent \textbf{Human Feedback Type. }
We categorize human feedback as \textit{evaluative}, \textit{corrective}, \textit{guidance}, and \textit{implicit} feedback. \textbf{\textit{(1) Evaluative Feedback}} provides an assessment of the agent’s output quality, typically as preference ranking, scalar rating, or binary assessment. A prime example is preference ranking, where users compare agent outputs, forming the basis of Reinforcement Learning from Human Feedback (RLHF) \cite{chaudhari2024rlhf}. Alternatively, platforms like Uni-RLHF \cite{yuan2024unirlhf} support scalar ratings or binary assessments. \textbf{\textit{(2) Corrective Feedback}} offers direct edits or fixes to the agent’s behavior. For instance, the PRELUDE \cite{gao2024aligning} framework learns latent preferences from user edits made to agent-generated text. \textbf{\textit{(3) Guidance Feedback}} means the human proactively provides instructions, critiques, or demonstrations to shape the agent’s behavior. Agents like InteractGen \cite{suninteractgen}, AutoManual \cite{chen2024automanual} can be bootstrapped using initial demonstrations, while methods like Self-Refine \cite{choudhury2025better} employ iterative critiques and refinements to improve outputs. \textbf{\textit{(4) Implicit Feedback}} is inferred by the agent observing user actions or control signals, rather than explicitly stated or direct output modifications. For example, an agent might learn user priorities by observing how a user adjusts control sliders in a system like VeriPlan \cite{Lee2025VeriPlanIF}, or infer preferences by analyzing user behaviors like clicks and purchases in frameworks such as AgentA/B \cite{wang2025agenta}. This contrasts with corrective feedback, where the user directly edits the output; here, the agent interprets the user's independent actions or control choices. \\

\noindent \textbf{Human Feedback Granularity. }
Human feedback also varies in granularity, from coarse-grained, holistic judgments to fine-grained, segment-level critiques. \textbf{\textit{(1) Coarse-grained/Holistic feedback}} provides a single assessment for the entire agent output. Standard RLHF often relies on holistic preferences between complete responses, which simplifies feedback collection but struggles with credit assignment in complex tasks. \textbf{\textit{(2) Fine-grained/Segment-Level Feedback}} by contrast, targets specific parts (e.g., sentences, paragraphs, code blocks). 
This is crucial in environments like ConvCodeWorld \cite{han2025convcodeworld}, where feedback pertains to specific conversational turns or generated code segments, or in annotation tasks like PDFChatAnnotator \cite{10.1145/3640543.3645174}, where feedback applies to specific annotations or parts of the document.
This finer granularity provides more precise learning signals, crucial for debugging complex behaviors. \\


\noindent \textbf{Human Feedback Phase. }
Human feedback can be incorporated at different phases of the LLM-agent pipeline \cite{wang2025m3hf}. \textbf{\textit{(1) Initial Setup \& Goal Definition}} occurs before task execution, configuring the agent system and defining goals, such as setting coordination strategies (AgentCoord \cite{pan2024agentcoord}) or critiquing plans before execution (Ask-before-Plan \cite{zhang-etal-2024-ask}). \textbf{\textit{(2) During Task Execution}} involves online, interactive feedback while the agent is actively performing the task, enabling real-time adaptation. Examples include interactive instruction editing (InstructEdit \cite{wang2023instructedit}), mid-task refinements (Mutual Theory of Mind \cite{Zhang2024MutualTO}, Collaborative Gym \cite{shao2024collaborative}), or online interventions (HG-DAgger \cite{kelly2019hg}, InterruptBench \cite{zou2026users}). \textbf{\textit{(3) Post-Task Evaluation \& Refinement}} happens after task completion to assess outcomes and provide feedback 
for future use. Frameworks like MAIH \citep{wang2024safe} and EmoAgent \citep{qiu2025emoagent} apply feedback loops after initial generation for benchmarking or offline learning, while AdaPlanner \cite{sun2023adaplanner} archives successful plans post-task as skills for future use. 
Table \ref{tab:feedback_dimensions} summarizes different dimensions of human feedback, key characteristics, and example work.

\subsection{Human-Agent Interaction Types}
\label{sec:interaction types}

\begin{figure}[!t]
  \centering
  \includegraphics[width=\columnwidth]{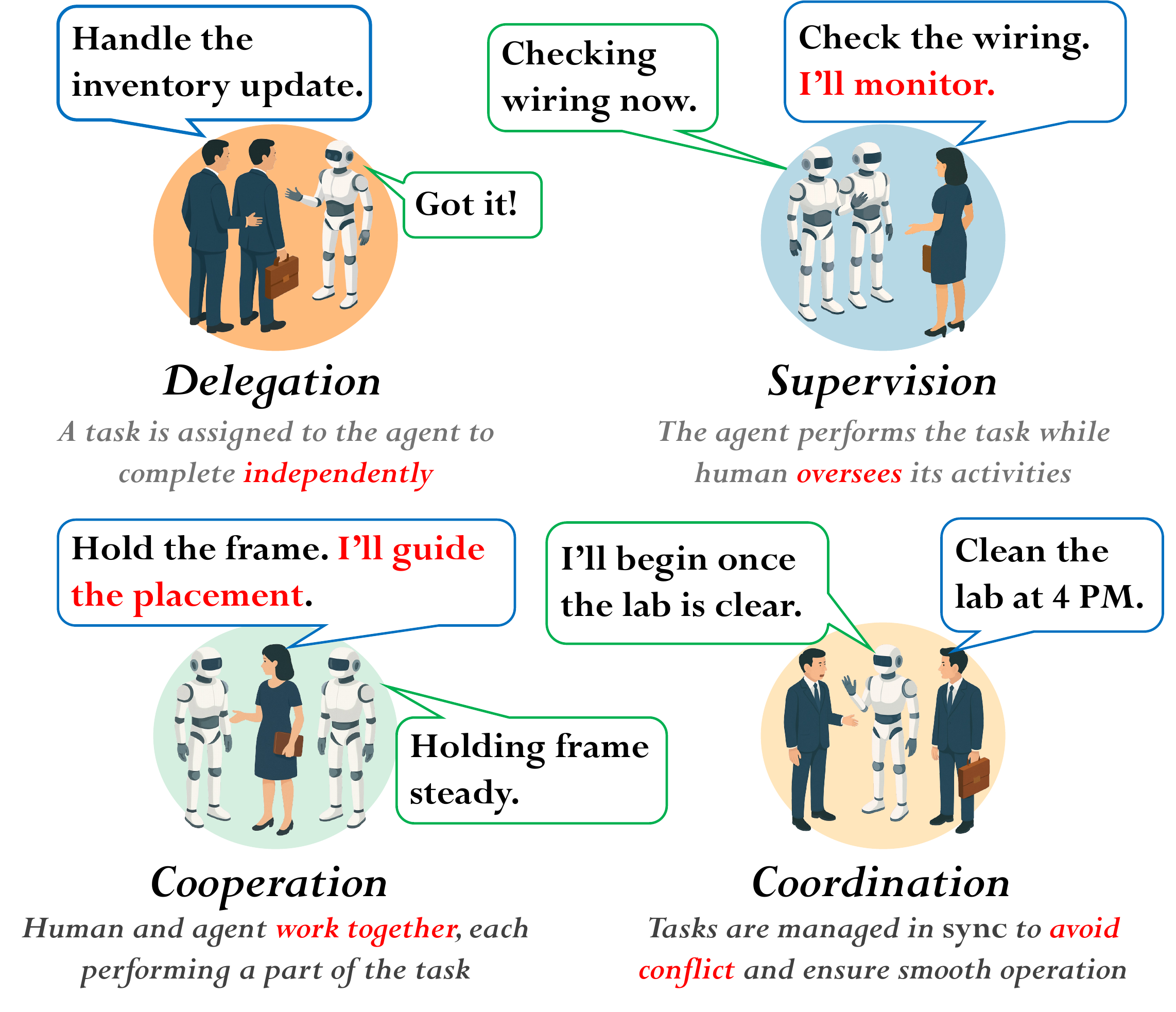}
  \caption{The subtype of the collaboration between humans and LLM-based agents.}
  \label{fig:subtype of collaboration}
\end{figure}

Interaction types define how individuals communicate, exchange information, and take actions with one another. In LLM-HAS, interactions tend to be more dynamic and complex compared to multi-agent systems. This complexity arises from the various roles and responsibilities assigned to both human agents and those based on LLMs, necessitating a finer-grained framework to describe their collaborative behaviors. The following categorization highlights the three key interaction types: \textbf{Collaboration}, \textbf{Competition}, and \textbf{Coopetition}. 

\subsubsection{Collaboration}
\label{sec:collaboration}
Collaborations are by far the most common interaction and foundational interaction, which involve humans and LLM-based agents working together to achieve a common goal. This partnership combines human creativity and contextual understanding with LLM-based agents to address challenges and improve the efficiency and quality of results \citep{vats2024survey, du-etal-2024-llms, sun2025symbioticrag,tang2026vividoc,chen2025embracing}. Depending on the type of collaboration considered, it can be categorized into four main fine-grained subtypes (Figure \ref{fig:subtype of collaboration}): 
\textit{\textbf{(1) Delegation \& Direct Command}} \citep{kiewiet1991logic}, 
\textit{\textbf{(2) Supervision}} \citep{loganbill1982supervision}, 
\textit{\textbf{(3) Cooperation}} \citep{rand2013human}, and
\textit{\textbf{(4) Coordination}} \citep{turvey1990coordination}. \\

\noindent \textbf{Delegation \& Direct Command.} 
In this interaction modality, a controlling party, usually a human, assigns explicit tasks to the LLM-based agent by providing clear and direct instructions. The agent is expected to execute these directives autonomously or on behalf of humans, ensuring that responsibilities are well-defined and actions align with the system's overarching objectives. Unlike supervision, where strategies can be dynamically adjusted in response to new situations, delegation involves providing instructions upfront. This means the agent follows a predetermined set of tasks rather than adapting to the situation. For instance, an investor specifies their risk preference to the agent executing the investment strategy like FineArena \citep{xu2025finarena}, or a driver utters the command to LLM-based agent like Drive as you Speak \citep{cui2024drive}. \\

\noindent \textbf{Supervision. } 
Supervision is the process by which one party, usually a human operator, oversees, monitors, and guides the actions of an LLM-based agent. This involves real time evaluation and intervention to ensure the agent’s output aligns with established goals and quality standards. Supervision also encompasses setting alert thresholds and providing corrective inputs when deviations occur. By maintaining a continuous feedback loop between the human and the agent, supervision helps calibrate agent behavior, catch and mitigate errors before they propagate and build confidence in the system. It also enables agents to handle routine tasks with increasing independence. For instance, agents notify humans to verify alignment \citep{Liu2023LLMPoweredHL}, and teleoperators monitor the LLM-generated motion plans \citep{liu2023llm}. \\

\noindent \textbf{Cooperation.} 
Cooperation refers to the voluntary and joint efforts of multiple parties to achieve agreed-upon goals. This collaboration type combines the various efforts and outcomes of different individuals and LLM-based agents toward a common objective. It emphasizes collective commitment, mutual assistance, and the pooling of resources to attain a shared result, thereby fostering a collaborative problem-solving environment. For instance, the human robot coordination in household activities \cite{chang2024partnr}, the cooperative embodied language agent (CoELA) \citep{zhang2024building}, human designers collaborating with an LLM-based agent \citep{sharma2024investigatingagencyllmshumanai}. \\

\noindent \textbf{Coordination.} 
Coordination is the organized process of aligning and synchronizing the actions of humans and LLM-based agents to achieve a shared objective. Unlike cooperation, the key idea behind coordination is to avoid conflict and bias in both humans and LLM-based agents to reach the final goal. It involves clear communication, strategic planning, and the intentional division of tasks, ensuring that individual efforts are harmonized and effectively integrated to support common goals. 
For example, humans and agents work in a shared workspace to complete interdependent tasks \citep{Zhang2024MutualTO}, human–agent integration supports adaptive decision-making \citep{suninteractgen}, and the collaborative framework facilitates coordination between humans and agents \citep{pan2024agentcoord}.

\subsubsection{Competition}
\label{sec:competition}
Competition is a form of interaction where participants aim to achieve their own goals, which often conflict with the objectives of others. In the LLM-HAS, competition emerges when agents or humans seek to enhance their personal performance or obtain resources, even if it negatively impacts collective results. In addition, competition also necessitates effective balancing mechanisms, like performance regulation or conflict resolution strategies, to prevent unproductive behaviors and ensure that the overall goals of the system remain intact. For instance, the SOTOPIA framework simulates social behaviors between humans and LLM-based agents \citep{zhou2024sotopia}.

\subsubsection{Coopetition}
\label{sec:coopetition}
Coopetition is an interaction where cooperation and competition coexist at the same time. Within this interaction, participants collaborate on shared tasks or mutual goals while also seeking to outdo each other to improve their own performance or gain extra advantages. In terms of the LLM-HAS, this dual aspect implies that agents and human may join forces to address complex issues while competing in specific domains such as efficiency or precision. This approach not only combines the strengths of both collaboration and competition, but also fosters innovation driven by competitive incentives while also reaping the benefits of cooperative synergy. Successfully managing coopetition typically requires mechanisms for building trust and adaptable strategies that reconcile collective advantages with personal aspirations, which is a challenge for the LLM-HAS. For example, humans and agents play the prisoner's dilemma in the shared workspace \citep{jiang2025experimentalexplorationinvestigatingcooperative}.

\definecolor{headercolor}{RGB}{90, 78, 140}     
\definecolor{strategy}{RGB}{245, 201, 211}      
\definecolor{strategysub}{RGB}{252, 228, 233}   
\definecolor{sync}{RGB}{200, 230, 209}          
\definecolor{syncsub}{RGB}{230, 245, 236}       
\definecolor{syncText}{RGB}{34,154,140} 
\definecolor{strategy}{RGB}{245, 201, 211}
\definecolor{strategyText}{RGB}{183,61,91}

\definecolor{elegantred}{RGB}{247,228,230}   

\definecolor{elegantmint}{RGB}{218,236,229}  
\definecolor{orchestClr}{RGB}{255,196,141} 

\colorlet{elegantredLight}{elegantred!55}    
\colorlet{elegantmintLight}{elegantmint!55}  
\begin{table}[!tp]
\centering
\renewcommand{\arraystretch}{1.4}
\setlength{\tabcolsep}{3pt}
\fontsize{8.5pt}{10pt}\selectfont

\begin{tabularx}{\linewidth}{%
    >{\raggedright\arraybackslash}p{2.1 cm}
    >{\raggedright\arraybackslash}X}
\toprule
\rowcolor{orchestClr}
\color{black}\textbf{Orchestration Paradigm} & \color{black}\textbf{Description} \\
\midrule

\rowcolor{elegantred}
\textbf{Task Strategy} & \textbf{\textcolor{strategyText}{What order and grouping}} of tasks do participants perform? \\

\rowcolor{elegantredLight}\textit{One‑by‑One} & \textbf{Actors take turns} (e.g., human plans $\rightarrow$ agent executes $\rightarrow$ human reviews $\rightarrow$ agent refines). \\

\rowcolor{elegantredLight}\textit{Simultaneous} & \textbf{Actors work in parallel} (e.g., agent streams partial suggestions while human types). \\

\rowcolor{elegantmint}\textbf{Temporal Synchronization} & \textbf{\textcolor{syncText}{When and how tightly}} do actors’ steps need to align in time? \\

\rowcolor{elegantmintLight}\textit{Synchronous} & (1) \textbf{Real-time interaction}: Humans and agents communicate simultaneously or in immediate sequence; (2) \textbf{Immediate response}: Participants expect or require prompt feedback. (e.g., live chat session, real-time voice assistant). \\

\rowcolor{elegantmintLight}\textit{Asynchronous} & (1) \textbf{Delayed interaction}: Communication occurs without the expectation of immediate responses; (2) \textbf{Flexible timing}: Participants can respond at their convenience. (e.g., email queues, human leaves comments, agent processes offline). \\

\bottomrule
\end{tabularx}

\caption{Orchestration paradigms in LLM-based human–agent systems encompass two orthogonal dimensions: task strategy, which can be one-by-one or simultaneous, and temporal synchronization, which can be synchronous or asynchronous.}
\label{tab:orch_paradigms}
\vspace{-3mm}
\end{table}

\subsection{Orchestration Paradigm}
\label{sec:orchestration paradigm}

The orchestration paradigm in LLM-HAS refers to \textit{how} tasks and interactions are managed between humans and agents, covering two dimensions in our survey: \textbf{Task Strategy} (\textit{ordering}) and \textbf{Temporal Synchronization} (\textit{timing}). Table \ref{tab:orch_paradigms} summarizes the two dimensions of the orchestration paradigm.

\subsubsection{Task Strategy}
\label{sec:strategy}
In LLM-HAS, the chosen task strategy, defined by the order and grouping of tasks performed by humans and agents, significantly impacts task execution effectiveness and efficiency. These strategies can typically be categorized into \textit{one-by-one} and \textit{simultaneous} paradigms.\\

\noindent \textbf{One-by-One. } 
The one-by-one strategy requires participants (humans and LLM-based agents) to perform tasks sequentially, taking clearly defined turns. For example, a human first outlines a plan, the agent then executes the task, the human subsequently reviews the output, and finally, the agent refines its work based on feedback \cite{liu2024enhancing, zhou2025sweetrltrainingmultiturnllm}. Such sequential interaction helps maintain a clear order of execution and reduces the complexity associated with concurrent task management. However, this rigidity may limit overall efficiency and flexibility, especially in dynamic scenarios requiring parallel processing or rapid interaction cycles \cite{bansal2024challenges, guo2024embodiedllmagentslearn}.\\

\noindent \textbf{Simultaneous. } 
Simultaneous strategy describes an interaction pattern in which LLM-based agents and humans respond concurrently in real time. Compared to the one-by-one strategy, the simultaneous approach more closely mirrors real-world conditions encountered in many simulation tasks \cite{wang2025m3hf, zhang2025leveraging}.
However, this strategy demands sophisticated mechanisms to handle latency mitigation and seamless coordination between participants.

\subsubsection{Temporal Synchronization}
Temporal synchronization in LLM-HAS refers to the timing and coordination of interactions between humans and agents. It significantly influences system responsiveness, user experience, and task efficiency. It can be broadly categorized into two modes: \textit{synchronous} and \textit{asynchronous}.\\

\noindent \textbf{Synchronous. } 
Synchronous interaction involves real-time interactions where humans and agents engage simultaneously. Immediate response is expected, facilitating dynamic exchanges. Examples include live chat sessions, real-time voice assistants (e.g., Siri, Alexa), and collaborative decision-making scenarios \cite{Zhang2024MutualTO, Liu2023LLMPoweredHL}. This mode is advantageous for tasks requiring rapid responses, immediate clarification, or real-time collaboration \cite{mehta-etal-2024-improving, han2025convcodeworld}.\\

\noindent \textbf{Asynchronous. }
In contrast, asynchronous interaction occurs without the necessity for immediate responses. Participants can engage at their convenience, allowing for flexibility in communication. Examples include email exchanges, message queues, ticket-based support systems, and task assignments where agents process and report outcomes independently \cite{shao2024collaborative, zhang2025leveraging}. Asynchronous communication is beneficial for complex issues that require thoughtful analysis or when participants are in different time zones \cite{suninteractgen, sun2024assistantx}.

\subsection{Communication}


\label{sec:communication}
In LLM-HAS, communication serves as the fundamental mechanism defining the transmission, reception, and transformation of information between humans and LLM-based agents. It focuses specifically on how \textit{information flows} across participants to support effective interaction and mutual understanding.
Unlike LLM-based multi-agent systems~\cite{yan2025beyond}, human-agent systems introduce a unique dimension (i.e., flexible, and cognitively diverse human participation). 
This leads to a broader and more complex communication landscape, encompassing both human-to-agent and agent-to-agent exchanges, each influenced by human interpretability, feedback style, and interaction latency.

To systematically analyze communication behavior in such systems, we propose a two-dimensional taxonomy that captures the communication behavior characteristics of humans and agents from macro-structures to micro-interaction rules. Specifically, we divide this section into the following parts: \textbf{Communication Structure}, which describes the macro-level organization of information channels, and \textbf{Communication Mode}, which characterizes the micro-level methods of message exchange.

\subsubsection{Communication Structure}
Communication structure refers to the organizational structure of agents, including both humans and agents, in LLM-HAS. 
It determines how information flows at the macro level and shapes the rules of interaction at the micro level. 
While originally developed for LLM-based multi-agent environments~\cite{guo2024large}, these structures have been effectively adapted to human-agent scenarios by treating humans as specialized agents. 
In such systems, the communication structure not only governs the efficiency of information exchange but also significantly impacts the system’s adaptability, scalability, and robustness to human variability. We categorize the representative structures into three types: \textbf{Centralized}, \textbf{Decentralized}, and \textbf{Hierarchical}.

In \textbf{Centralized} structure, one primary agent or a group of core agents acts as a central node to coordinate all communications within the system. This central agent manages interactions among other agents, simplifying coordination and minimizing conflicts~\cite{cui2024drive}. 
\textbf{Decentralized} structure employs peer-to-peer communication, enabling direct interactions among agents without centralized control. Agents autonomously manage their communications based on systemic information, enhancing system flexibility, adaptability, and robustness~\cite{shao2024collaborative, 10.5555/3618408.3618748}. 
In addition, \textbf{Hierarchical} structure organizes agents into clearly defined levels, assigning distinct roles and responsibilities according to their position within the hierarchy~\cite{Liu2023LLMPoweredHL,pan2024human}. 
High-level agents typically fulfill managerial or strategic roles, providing overarching guidance and supervision, while lower-level agents perform specialized tasks and execute detailed operations. 

\subsubsection{Communication Mode}
Communication mode defines the manner through which humans and agents exchange information within LLM-HAS. Specifically, communication mode describes the methods employed by participants to transmit, acquire, and utilize information, critically shaping interaction efficiency and the overall performance of the system. Broadly, communication modes can be categorized into three primary approaches: \textbf{Conversation}, \textbf{Observation}, and \textbf{Shared Message Pool}. \\

\noindent \textbf{Conversation.} The conversation-based mode is currently the most prevalent and intuitive approach in LLM-HAS, wherein agents and humans directly engage through natural language dialogues. This interaction format typically utilizes conversational interfaces that allow iterative exchanges, questions, clarifications, and dynamic responses, facilitating efficient collaboration and mutual understanding~\cite{shao2024collaborative}. For instance, conversational LLM agents can assist users by answering queries, explaining complex concepts, or collaboratively solving reasoning tasks through iterative dialogues~\cite{wang2024mint}. While intuitive and flexible, conversational interactions rely significantly on the communicative clarity and dialogue management capabilities of LLM agents. \\

\begin{table*}[!th]
\centering
\begin{threeparttable}
\small
\begin{tabularx}{\textwidth}{c l X l}
\toprule
\textbf{Level} & \textbf{Name} & \textbf{Description} & \textbf{Agent Role} \\
\hline
A1 & Full Automation & Agent handles task entirely without human involvement & Automation \\
\hline
A2 & Minimal Human Input & Agent needs human input only at key points (e.g., spot-checking or exception handling) & Automation \\
\hline
A3 & Equal Partnership & Human and agent collaborate closely, outperforming either alone (e.g., planning/analysis tasks requiring iterative back-and-forth) & Augmentation \\
\hline
A4 & Agent-Assisted & Agent requires substantial human input to complete task & Augmentation \\
\hline
A5 & Human-Driven & Task fully relies on continuous human involvement & Augmentation \\
\bottomrule
\end{tabularx}
\end{threeparttable}
\caption{Human Agency Scale~\cite{shao2025future}, a system-level framework that quantifies the desired or required level of human involvement in LLM-HAS. \textbf{Automation:} The agent takes primary responsibility for task execution with minimal human oversight. \textbf{Augmentation:} The human retains meaningful involvement with Agent providing collaborative support. The scale spans five levels (A1--A5), ranging from full automation to human-driven workflows. Detailed discussions of the motivation and each level are provided in Section~\ref{sec:human_agency_scale} and Appendix~\ref{app:HAAS}.}
\label{tab:haas_levels}
\end{table*}

\noindent \textbf{Observation.} In the observation-based communication mode, agents acquire information implicitly by observing participants behaviors, decisions, or interactions within their environment, rather than through explicit verbal communication. This mode leverages indirect signals, including user actions, feedback cues, or behavioral traces, to infer intentions, preferences, or states~\cite{lu2024turn}. For example, an LLM-driven tutoring system may adaptively provide targeted instructions by continuously observing student problem-solving behaviors without explicit verbal queries~\cite{pan2024human}. However, relying solely on observational signals can introduce ambiguity, potentially impacting inference accuracy unless complemented by robust inferential mechanisms. \\

\noindent \textbf{Message Pool.} The shared message pool mode involves agents and humans exchanging information through a common information repository. Participants publish messages or data into a message pool, subscribing and retrieving relevant messages based on specific interests or tasks~\cite{sun2024assistantx}. This approach significantly simplifies direct agent-to-agent or human-to-agent interactions, reduces communication complexity, and enhances information management efficiency. A prominent example includes the MetaGPT framework~\cite{hong2023metagpt}, where LLM-based agents collaboratively retrieve information dynamically from a shared message pool, streamlining cooperation and information dissemination. Despite these advantages, shared message pools must carefully manage access control to avoid information conflicts or inefficient retrieval.

\subsection{Human Agency Scale}
\label{sec:human_agency_scale}
The five components discussed above collectively characterize \textit{\textbf{how humans and agents collaborate or interact.}} However, they do not directly address a fundamental question: \textit{\textbf{to what extent should humans be involved in task completion?}} Different tasks and application contexts call for varying degrees of human participation, ranging from full automation to essential human involvement. Drawing on recent work that examines worker preferences and technological capabilities across occupational tasks~\cite{shao2025future}, we introduce the Human Agency Scale, a system-level framework that quantifies the desired or required level of human involvement in LLM-HAS~\cite{shao2025future}.
This scale defines five levels based on the degree of human involvement required for effective task completion: \textit{A1:Full Automation}, \textit{A2:Minimal Human Input}, \textit{A3:Equal Partnership}, \textit{A4:Agent-Assisted} and \textit{A5:Human-Driven}. Levels A1–A2 correspond to \textbf{automation}, where agent replaces human effort, while A3–A5 represent \textbf{augmentation}, where agent enhances human capabilities. A detailed discussion of each level is provided in Appendix~\ref{app:HAAS}.



\section{Implementation Strategies}
\label{sec:implement_strategies}

This section compares major implementation strategies adopted in LLM-based human-agent systems. Specifically, we include three widely-used categories: 1) Prompting-based methods, (2) Supervised Fine-Tuning (SFT)-based methods, and (3) Reinforcement Learning (RL)-based methods. For each category, we summarize representative methods and analyze their strengths and limitations.\\ 

\noindent \textbf{Prompting-based collaboration} remains the most widely adopted strategy due to its flexibility, easy implementation and minimal training overhead. Recent work demonstrates that carefully structured prompts can elicit sophisticated collaborative behaviors, such as proactive clarification, shared planning, and theory-of-mind reasoning. Systems like MToM \cite{Zhang2024MutualTO} and Collaborative Gym \cite{shao2024collaborative} show that explicit role, belief, or goal modeling in prompts enables agents to anticipate user intent and adapt responses accordingly. Interactive benchmarks and interfaces, such as RECODE-H \cite{miao2025recode}, Magentic-UI \cite{Mozannar2025MagenticUITH}, InterruptBench \cite{zou2026users}, and ARIA \cite{he-etal-2025-enabling}, further illustrate how real-time human feedback (e.g., critiques, corrections, or preferences) can be injected into the agent loop at inference time to guide task execution and self-improvement. Analyses of real-world usage, such as PATHs \cite{mysore-etal-2025-prototypical}, reveal recurring human–AI collaboration patterns that prompting can exploit without modifying model parameters. However, despite their agility, prompting-based methods are often brittle: behaviors are sensitive to prompt design, have limited controllability, and struggle to accumulate learning across sessions.\\

\noindent \textbf{Supervised fine-tuning (SFT)} addresses these limitations by converting human interaction traces, such as edits, revisions, or clarifications, into persistent behavioral improvements. Works like PRELUDE \cite{gao2024aligning} and XtraGPT \cite{Chen2025XtraGPTCA} demonstrate how user edits can be treated as supervision signals, allowing agents to learn latent user preferences or revision strategies beyond single-turn prompting. Hybrid systems, such as Ask-before-Plan \cite{zhang-etal-2024-ask} and CollabLLM \cite{wu2025collabllm}, combine prompting with SFT to balance adaptability and stability, enabling agents to proactively ask questions while grounding behavior in learned collaboration policies. Compared to prompting, SFT yields more consistent agent behavior and stronger performance on specific tasks, but it incurs higher data and engineering costs and remains constrained by the coverage and bias of collected interaction data.\\

\begin{table*}[!t]
  \centering
  \fontsize{8}{9}\selectfont
  \renewcommand{\arraystretch}{1.1} 
  \begin{tabular}{@{}lllc@{}}
    \toprule
    \textbf{Domain} & \textbf{Datasets \& Benchmarks} & \textbf{Proposed or Used by} & \textbf{Data Link} \\
    \hline
    \multirow{8}{*}{Embodied AI}
      & TaPA               & \paper{TaPA}{wu2023embodied}                           & \codelink{https://github.com/Gary3410/TaPA} \\
      & EmboInteract       & \paper{InteractGen}{suninteractgen}                    & \textit{--} \\
      & AssistantX         & \paper{AssistantX}{sun2024assistantx}                  & \textit{--} \\
      & IGLU Multi‑Turn    & \paper{Help Feedback}{mehta-etal-2024-improving}       & \codelink{https://github.com/microsoft/iglu-datasets} \\
      & PARTNR             & \paper{PARTNR}{chang2024partnr}                        & \codelink{https://github.com/facebookresearch/partnr-planner/tree/main/} \\
      & MINT               & \paper{MINT}{wang2024mint}                             & \codelink{https://github.com/xingyaoww/mint-bench} \\
      & C‑WAH              & \paper{REVECA}{seo2025reveca}                          & \codelink{https://github.com/UMass-Embodied-AGI/CoELA} 
      \\
      & HSRI              & \paper{HSRI}{lee2025human}                          & \textit{--} \\
    \hline
    \multirow{7}{*}{Conversational Systems}
      & WEBLINX                   & \paper{WebLINX}{lu2024weblinx}                        & \textit{--} \\
      & Ask‑before‑Plan           & \paper{Ask Before Plan}{zhang-etal-2024-ask}          & \codelink{https://github.com/magicgh/Ask-before-Plan} \\
      & Agency Dialogue   & \paper{Agency Task}{sharma2024investigatingagencyllmshumanai} & \textit{--} \\
      & WildSeek                  & \paper{Co‑STORM}{jiang-etal-2024-unknown}             & \codelink{https://github.com/stanford-oval/storm?tab=readme-ov-file} \\
      & MINT                      & \paper{MINT}{wang2024mint}                            & \codelink{https://github.com/xingyaoww/mint-bench} \\
      & HOTPOTQA                  & \paper{ReHAC}{feng-etal-2024-large}                          & \codelink{https://hotpotqa.github.io/} \\
      & StrategyQA                & \paper{ReHAC}{feng-etal-2024-large}                          & \codelink{https://github.com/eladsegal/strategyqa} \\
    \hline
    \multirow{5}{*}{Software Development}
      & MINT                      & \paper{MINT}{wang2024mint}                            & \codelink{https://github.com/xingyaoww/mint-bench} \\
      & InterCode                 & \paper{ReHAC}{feng-etal-2024-large}                          & \codelink{https://github.com/princeton-nlp/intercode} \\
      & ColBench                  & \paper{SWEET‑RL}{zhou2025sweetrltrainingmultiturnllm} & \codelink{https://huggingface.co/datasets/facebook/collaborative_agent_bench} \\
      & ConvCodeWorld             & \paper{ConvCodeWorld}{han2025convcodeworld}           & \codelink{https://github.com/stovecat/convcodeworld} \\
      & ConvCodeBench             & \paper{ConvCodeWorld}{han2025convcodeworld}           & \codelink{https://github.com/stovecat/convcodeworld} \\
      & RECODE-H                      & \paper{RECODE-H}{miao2025recode}                            & \codelink{https://github.com/ChunyuMiao98/RECODE-H} \\
    \hline
    \multirow{2}{*}{Gaming}
      & CuisineWorld              & \paper{MindAgent}{gong2023mindagent}                  & \codelink{https://github.com/mindagent/mindagent} \\
      & MineWorld                 & \paper{MineWorld}{guo2025mineworld}                   & \codelink{https://github.com/microsoft/MineWorld} \\
    \hline
    \multirow{2}{*}{Healthcare}
      & EmoEval  & \paper{EmoAgent}{qiu2025emoagent}                     & \codelink{https://github.com/1akaman/EmoAgent} \\
      & GenoTEX  & \paper{GenoMAS}{Liu2025GenoMASAM}                     & \codelink{https://github.com/Liu-Hy/GenoTEX} \\
    \hline
    \multirow{2}{*}{Retail}
      & $\tau$2-Bench  & \paper{$\tau$2-Bench}{Barres20252BenchEC}                     & \codelink{https://github.com/sierra-research/tau2-bench} \\
      & $\tau$-Bench  & \paper{$\tau$-Bench}{yao2025taubench}                     & \codelink{https://github.com/sierra-research/tau-bench} \\
    \hline
    \multirow{3}{*}{Travel}
      & UserBench  & \paper{UserBench}{Qian2025UserBenchAI}                     & \codelink{https://github.com/SalesforceAIResearch/UserBench} \\    
      & $\tau$2-Bench  & \paper{$\tau$2-Bench}{Barres20252BenchEC}                     & \codelink{https://github.com/sierra-research/tau2-bench} \\
      & $\tau$-Bench  & \paper{$\tau$-Bench}{yao2025taubench}                     & \codelink{https://github.com/sierra-research/tau-bench} \\
    \hline
    Finance
      & FinArena‑Low‑Cost  & \paper{FineArena}{xu2025finarena}                     & \codelink{https://huggingface.co/datasets/Illogicaler/FinArena-low-cost-dataset} \\
    \hline
    Web Navigation \& Computer Use
      & InterruptBench  & \paper{InterruptBench}{zou2026users}                     & \codelink{https://github.com/HenryPengZou/InterruptBench} \\
    \bottomrule
  \end{tabular}
  \caption{Datasets and Benchmarks across various domains.}
  \label{tab:benchmarks}
\end{table*}

\noindent \textbf{Reinforcement learning (RL)} formulates human–agent interaction as a sequential decision-making problem with explicit reward objectives. Recent RL-based work, such as UserRL \cite{Qian2025UserRLTI}, MUA-RL \cite{Zhao2025MUARLMU}, SWEET-RL \cite{zhou2025sweetrltrainingmultiturnllm}, and ReHAC \cite{feng-etal-2024-large}, optimizes agents for proactive help-seeking, tool use, and multi-turn coordination under delayed rewards. Interactive environments like UserBench \cite{Qian2025UserBenchAI} provide controlled testbeds for evaluating user-centric policies, moving beyond static benchmarks toward longitudinal interaction. Compared to prompting and SFT, RL enables agents to reason over long horizons and trade off immediate assistance against future user satisfaction. However, RL approaches face challenges in reward specification, sample efficiency, and training stability. Thus, many recent works \cite{Zhao2025MUARLMU, Qian2025UserRLTI} adopt hybrid pipelines that bootstrap RL from prompting or SFT, suggesting that effective human–agent collaboration arises from complementary learning signals rather than a single paradigm.

\section{Applications}

\label{sec:application_resource}



A diverse range of applications has emerged for LLM-HAS. We elaborate on the five most frequent domains below and summarize corresponding datasets and benchmarks in Table \ref{tab:benchmarks}. With new applications appearing almost weekly in this fast-growing field, we maintain a \href{https://github.com/HenryPengZou/Awesome-Human-Agent-Collaboration-Interaction-Systems}{GitHub repository} to track recent developments. \\

\label{sec:application}
\noindent \textbf{\domainembodied.}
Applications in Embodied AI involve various aspects of dynamic and complex real-world tasks, benefiting from valuable human feedback and interactions in LLM-HAS. \citet{10141597} explores incorporating LLMs in human-robotic collaboration assembly tasks, allowing seamless communication between robots and humans and increasing trust in human operators. To address the challenges of false planning due to suboptimal environment changes, \citet{seo2025reveca} proposes REVECA to enable efficient memory management and optimal planning. Additionally, \citet{tanneberg2024help} extends the agents' collaboration with a group of humans via Attentive Support, enabling agents' ability to remain silent to not disturb the group if desired. \\

\noindent \textbf{\domainsoftware.}
The inherently collaborative nature of software development makes human-agent collaboration vital to improve development efficiency \cite{lu2024turn, han2025convcodeworld, zhou2025sweetrltrainingmultiturnllm}. \citet{feng-etal-2024-large} introduces ReHAC framework, wherein agents are trained to determine the optimal stages for human intervention within the problem-solving process, offering improved generalizability over the traditional heuristic-based approaches. Building on this direction,~\citet{zhou2025sweetrltrainingmultiturnllm,han2025convcodeworld,wang2024mint} investigate a broader spectrum of human feedback types via multi-turn human-agent interactions. These approaches incorporate carefully designed optimization objectives to effectively capture more diverse and nuanced interactions between humans and agents. \\

\noindent \textbf{\domainconversation.}
In conversational systems, due to the frequent presence of ambiguity and lack of necessary information that agents cannot reliably infer, such as login credentials and payment details, effective human-agent collaboration constitutes a critical component of the system. \citet{zhang-etal-2024-ask} introduces Proactive Agent Planning, wherein agents are trained to predict classification needs based on the user-agent conversational interactions and current environment, thereby leading to improved reasoning efficacy. \citet{10.1145/3491102.3517582} introduces Chaining the LLM to improve the quality of task outcomes and enhance the transparency and controllability of the conversational systems. \\

\noindent \textbf{\domaingaming. } 
LLM-HAS are naturally well-suited to simulated gaming environments due to their dynamicity and sophistication. Proper human-agent interactions have been shown to enhance humans' experience, satisfaction and understanding of both the environment and agents \cite{gong2023mindagent,gao2024enhancing}. Collaborative interactions also contribute to improved agents' task performance and decision-making capabilities.
 For instance, MindAgent framework \cite{gong2023mindagent} illustrates the efficacy of human-agent collaboration through measurable improvements in task outcomes when humans and agents work together. \citet{mehta-etal-2024-improving} demonstrates agents achieve improved outcomes when interacting with humans via autonomous confusion detection and clarification questions and inquiries. \citet{ait2024towards} introduces Meta-Command Communication-based framework to enable effective human-agent collaboration. To address challenges related to execution latency while maintaining strong reasoning capabilities, \citet{liu2023llm} proposes Hierarchical Language Agent that promotes faster responses, stronger cooperation, and more consistent language communications. \\

\noindent \textbf{\domainfinance. }
Given the complexity of stock markets and financial systems, where investors' strategies and risk preferences are critical determinants of outcomes, human-agent collaboration is increasingly recognized as a valuable paradigm. FinArena~\cite{xu2025finarena} demonstrates the potential of integrating experienced investors with advanced AI agents to support stock prediction tasks. This collaborative framework has been shown to improve investment performance, yielding competitive annualized returns and Sharpe ratios~\cite{xu2025finarena}.

\appsection{Implementation Tools and Resources}
\label{sec:implementation}

\appsubsection{Human-Agent Framework}
This section provides a detailed introduction to three representative open-source LLM-HAS frameworks: Collaborative Gym \cite{shao2024collaborative}, COWPILOT \cite{huq2025cowpilotframeworkautonomoushumanagent}, and DPT-Agent \cite{zhang2025leveraging}. 
They differ in key configuration aspects, including environment settings, interaction types, orchestration paradigms, and communication strategies.
Specifically, 
\textbf{Collaborative Gym} \cite{shao2024collaborative} facilitates asynchronous interactions among humans, agents, and task environments, supporting various simulated and real-world tasks such as travel planning, data analysis, and academic writing. It emphasizes flexible, real-time collaboration and evaluates both outcomes and interaction quality, making it a robust tool for studying human-agent dynamics.
\textbf{COWPILOT} \cite{huq2025cowpilotframeworkautonomoushumanagent} provides a framework for human-agent collaborative web navigation through a Chrome extension. It employs a "Suggest-then-Execute" model under human supervision, allowing dynamic interventions to enhance task completion rates and reduce human workload. It effectively demonstrates how human intervention can significantly improve agent performance.
\textbf{DPT-Agent} \cite{zhang2025leveraging} applies Dual Process Theory (DPT) to enable real-time simultaneous human-agent interactions. It features intuitive, fast decision-making and deliberative reasoning components, employing Theory of Mind and asynchronous reflection to manage latency and adapt dynamically to human actions. This approach excels in environments requiring immediate and adaptive responses.

Other frameworks, such as \textbf{A2C}~\cite{tariq2025a2c}, \textbf{FinArena}~\cite{xu2025finarena}, and \textbf{human-robot collaboration framework}~\cite{liu2023llm}, also contribute significantly to specific domains like cybersecurity, financial forecasting, and robotic manipulation, respectively. These frameworks further demonstrate the diverse potential and adaptability of LLM-HAS.

\appsubsection{Datasets and Benchmarks}
We summarize the commonly used datasets and benchmarks for Large Language Model-based Human-Agent Systems in Table \ref{tab:benchmarks}. Diverse domains employ distinct methodologies for evaluating these systems, aligned closely with their unique application contexts. Within the domain of embodied AI, the primary approach involves simulated environments~\cite{suninteractgen, sun2024assistantx,mehta-etal-2024-improving}, designed to assess how effectively agents cooperate and execute tasks in dynamic, interactive scenarios. Another significant domain, Conversational Systems, encompasses applications such as question answering~\cite{feng-etal-2024-large}, website navigation~\cite{lu2024weblinx, levy2024st}, design decision assistance~\cite{sharma2024investigatingagencyllmshumanai}, and travel planning~\cite{zhang-etal-2024-ask}, adopting benchmarks that evaluate the ability of language models to function as user-aligned conversational assistants, ensuring interactions meet user expectations. Despite the extensive application coverage of current benchmarks, there remains a clear necessity for the development of more comprehensive and standardized benchmarking frameworks.

\section{Challenges and Opportunities}
\label{sec:challenge}




\label{sec:challenge_future}

In this section, we highlight some existing challenges and opportunities for LLM-HAS. \\

\noindent \textbf{Human Flexibility and Variability. } 
Human feedback varies widely in terms of role, timing, and style across various LLM-HAS. Humans are often subjective, influenced by their personalities, which means different individuals interacting with an LLM-HAS may lead to different outcomes and conclusions. This highlights the need and opportunity for i) thorough investigations or benchmarks on how varied human feedback affects entire systems, and ii) flexible frameworks that can support and adapt to diverse human feedback.
In addition, humans, regarded as a ''special agent'' in the LLM-HAS, are subject to fewer restrictions and evaluations than LLM-based agents. This limits how the LLM-HAS can be improved because the impedance may be on the human side instead of the agent. This concern remains and requires a refined strategy to define the strict, fine interaction rule and evaluation equally for both human and LLM-based agents. Also, many studies today substitute real human participants with LLM simulated human proxies, failing to capture human input's variety and unpredictability. For example, CollabLLM \citep{wu2025collabllm} employs a user simulator to mimic human interaction according to a predefined linguistic style. Nevertheless, the model still relies on fixed prompts to reproduce the requested actions, and its internal knowledge far exceeds that of an average human. As a result, the simulated conversation rarely involves extensive classification and verification steps \citep{yao2025taubench}, which are often expressed in imprecise language in the human perspective. In contrast, real users frequently produce grammatical errors or struggle to articulate their intentions clearly, behaviors that are rarely observed in LLM agents. The performance gap between humans and the simulated human remains unknown, potentially making the comparison incomparable. \\

\noindent \textbf{Mostly Agent-Centered Work.}  
In most LLM-HAS studies, guidance flows in a single direction, with humans evaluating agent outputs and providing corrective or evaluative feedback. Namely, the current studies are mostly agent-centered. This agent-centered framework relegates humans to passive evaluators and overlooks the potential for agents to proactively monitor and guide human actions, thereby undercutting bidirectional collaboration \citep{zhang2024building}. For example, ConvCodeWorld \citep{han2025convcodeworld} treats humans as scripted evaluators. Within the framework, LLM-simulated humans provide feedback logs to the agent, rather than empowering agents to observe and coach human coding actions dynamically. However, enabling agents to observe human actions, detect errors or inefficiencies, and offer timely suggestions can transform collaboration and reduce human effort by leveraging agent intelligence. When agents act as instructors by proposing alternative strategies, drawing attention to overlooked risks, and reinforcing effective practices as tasks unfold in real time, both humans and agents benefit. We believe that exploring human-centered LLM-HAS, or shifting toward an equalized LLM-HAS, will unlock the full promise of teamwork between humans and agents. \\

\noindent \textbf{Inadequate Evaluation Methodologies.} 
In existing evaluation frameworks for LLM-HAS, improvements focus primarily on agent accuracy and static benchmarks, which ignore the real burden placed on human collaborators~\cite{ma2025sphere}. People dedicate varying amounts of time, attention and cognitive effort depending on the type and frequency of feedback they must provide, yet no standard metric captures this human workload or its impact on overall efficiency. For example, frameworks like CoELA \citep{jiang-etal-2024-unknown} evaluate success purely by metrics such as transport rate improvements, yet ignore entirely the invisible coordination costs and cognitive load on humans \citep{qiu2025emoagent}. Evaluation methods should measure factors such as time spent offering feedback, feedback quality, frequency, and impact \citep{fragiadakis2024evaluating}, perceived mental workload and effort required to detect and correct errors, and they should cover every phase of the human agent collaboration from initial task assignment through post execution review, to systematically evaluate LLM-HAS. Evaluation methods should measure factors such as time spent offering feedback, perceived mental workload and effort required to detect and correct errors, and they should cover every phase of the human agent collaboration from initial task assignment through post execution review. As human expertise and LLM-based agent capabilities merge to deliver unprecedented performance, both uncertainty and variability grow. A new evaluation approach or set of metrics that systematically and comprehensively quantifies contributions and costs for both humans and agents is essential to ensure truly efficient collaboration.\\

\vspace{1mm}
\noindent \textbf{Unresolved Safety Vulnerabilities.} 
Most LLM-HAS works emphasize improving agent performance and have left safety, robustness and privacy underexplored in the context of human interaction~\cite{qiu2025emoagent}. As people and LLM-based agents collaborate in dynamic workflows, the risk of misaligned behavior, unexpected failures, or unintended disclosure of sensitive information grows. For example, the MetaGPT agent-centered framework \citep{hong2023metagpt}, while integrating task decomposition and communication, fails to integrate essential safety measures such as input sanitization, privacy-preserving data handling, and robust error-containment protocols. The MINT benchmark \citep{wang2024mint}, though it quantifies performance gains from multi-turn tool use and language feedback, omits any analysis of whether these interactive protocols might be exploited for code-injection attacks, data exfiltration, or other emergent safety failures. Humans engaging with these systems need clear safeguards around data sharing, error recovery protocols when agents behave unpredictably and privacy protections that cover every stage of the interaction. Robustness measures must ensure agents handle ambiguous or adversarial inputs without passing harm on to their human partners~\cite{glickman2025human}. Without studies that emphasize human experience in safety and privacy design, real-world deployments will struggle to gain trust or meet acceptable risk thresholds. Rigorous investigation of how safety, robustness and privacy shape human agent workflows from design through deployment is essential to build collaborations that are both effective and respectful of human needs. \\

\noindent \textbf{Applications and Beyond.} 
The potential of LLM-HAS extends well beyond current applications. Many opportunities remain to be explored in challenging domains such as healthcare, finance, scientific research, education, and so on \cite{luo2025large, guo2024large}. While fully autonomous LLM-based agent systems encounter difficulties in handling complex, long-term tasks and earning full trust in safety and reliability, the involvement of humans to provide additional information, feedback, and control allows LLM-HAS to greatly improve overall system performance and safety. This opens the door to impactful applications across a broad range of critical fields.



\section{Conclusion}
This paper presents a comprehensive review of LLM-based Human-Agent Systems. We introduce a structured taxonomy covering five core dimensions: environment and profiling, human feedback, interaction types, orchestration paradigms, and communication, and use it to classify and analyze existing research on LLM-HAS. We also summarize representative implementation frameworks, benchmark datasets, and evaluation metrics to support reproducibility and comparative analysis. Finally, we identify key challenges and unresolved issues in current LLM-HAS research. These issues remain major obstacles to the development of effective, adaptive, safe and trustworthy human-agent systems. We hope this review offers a comprehensive understanding of the LLM-HAS landscape and serves as a practical guide for future research.

\section*{Limitations}
Although we strive to include a wide range of representative works (e.g., ACL, EMNLP, NAACL, EACL, COLM, NeurIPS, ICLR, ICML, etc.), some relevant research may not be included, especially recent preprints or interdisciplinary research in fields such as cognitive science.

\section*{Acknowledgements}

This work is supported in part by NSF under grants III-2106758, and POSE-2346158.
Any opinions, findings, and conclusions expressed
here are those of the authors and do not necessarily reflect the views of NSF. We thank our reviewers for their insightful feedback and comments which helped improve the quality of our paper.

\bibliography{custom}

\newpage
\appendix
\label{sec:appendix}

\clearpage
\part{Appendix} 
\parttoc 











\appsection{Motivation: Why LLM-based Human-Agent Systems}  
\label{sec:appendix_motivation}  
Despite the rapid development of fully autonomous agents based on LLMs in recent years, such systems face persistent challenges in terms of reliability, complexity, and safety and ethical risks. Autonomous agents frequently generate incorrect or misleading information and often lack a true understanding of human goals or contextual nuances. These limitations suggest that full autonomy may not be suitable for many real-world applications\citep{mitchell2025fully, natarajan2025human, zou2025call} and highlight a critical but often overlooked insight: the indispensable role of human involvement.


Humans can provide important complementary capabilities such as disambiguation, domain-specific knowledge, feedback, corrections, and high-level supervision that are difficult for automated systems to replicate \citep{wang2024golf, dutta2024problem}. These factors have led to growing interest in a new class of systems explicitly designed for human-agent collaboration: \emph{\textbf{LLM-based Human-Agent Systems (LLM-HAS)}}. Rather than aiming to replace humans, LLM-HAS frameworks are presupposed with active human involvement, leveraging human expertise, supervision, and guidance to compensate for the limitations of the autonomy agent system.


By integrating human collaborators, LLM-HAS systems become more trustworthy, adaptable, and context-aware. In high-stakes fields such as healthcare or finance, collaborative systems are better able to handle complex and sensitive tasks than standalone agent systems. For example, the success of the "FTT" \citep{everett2025tool} (a hybrid team of human clinicians and agents) in the diagnostic reasoning challenge proves that human-agent collaboration can surpass both humans and agents alone. Human-agent collaboration allows building intelligent systems that are more robust, ethical, and value-aligned than either humans or agents could achieve alone.




\appsection{Evolution of Human-Agent Collaboration and Interaction Systems}
\label{sec:appendix_evolution}



The evolution of human-agent collaboration and interaction systems has been shaped by a series of major paradigm shifts. Early systems were grounded in rule-based and symbolic AI approaches, relying on predefined rules and handcrafted scripts to simulate human interactions. Iconic systems such as ELIZA \citep{weizenbaum1966eliza} and SHRDLU \citep{winograd1972understanding} enabled basic language interactions, but only within highly constrained environments. These agents acted as deterministic tools, offering little room for flexibility, adaptation, or learning from user behaviors. Interactions were largely one-directional, with humans issuing commands and agents executing tasks in a fixed, scripted manner.

The second major shift emerged with the rise of machine learning and data-driven NLP in the 2010s. Fueled by large annotated corpora and supervised learning techniques, agents began to exhibit more robust and flexible behavior, particularly in tasks such as speech recognition and dialogue generation \citep{young2013pomdp}. Commercial systems like Siri and Google Assistant exemplified this transition, allowing broader-domain conversations and more natural interactions. However, these systems remained primarily agent-centric: optimization efforts largely focused on improving task performance, with less attention given to human adaptivity or user-centered design. Users were often treated as passive input providers, with limited opportunities for active collaboration, personalization, or mutual learning \citep{madotto2019personalizing, liao2020questioning}.

In recent years, the advent of large language models (LLMs) has fundamentally transformed the landscape of human-agent collaboration. Systems powered by LLMs, such as ChatGPT and Claude, exhibit remarkable abilities in reasoning, co-creation, and open-ended problem solving. These agents move beyond reactive responses to actively engage in collaborative workflows with humans, supporting iterative refinement, clarification, and joint decision-making \citep{lou2025unraveling, yan2025passive}. This new paradigm emphasizes human-AI collaboration as a core design principle, spurring a growing focus on human-centered design, personalized adaptation, and interactive learning. As research shifts from optimizing isolated agent capabilities toward co-optimizing human-agent systems as integrated, adaptive teams, human adaptivity, transparency, and control are increasingly prioritized as central components of effective collaboration \cite{Qian2025UserBenchAI,sun2025training}.

\appsection{Human Agency Scale}
\label{app:HAAS}


The development of LLM-based human-agent systems raises a fundamental question: \textit{\textbf{how much human involvement is appropriate for a given task?}} Traditional perspectives on LLM-based Agents often adopt an "Agent-first" view, focusing primarily on the extent to which tasks can be automated. While useful for assessing technological capabilities, such perspectives do not adequately capture the human-centered considerations essential for responsible agent deployment, including human preferences, decision-making authority, and accountability~\cite{zou2025call}.
Drawing on recent work that examines worker preferences and technological capabilities across occupational tasks~\cite{shao2025future}, human agency scale provides a unified framework for quantifying the degree of human involvement required or desired in human-agent collaboration. This framework centers on \textbf{human agency} (i.e., the capacity for humans to exercise meaningful control, judgment, and decision-making authority within the system).

Human agency scale serves three key purposes in the context of LLM-HAS:
\begin{itemize}
\item \textbf{System Design:} Helping designers determine appropriate configurations of interaction types, feedback mechanisms, and orchestration paradigms based on target agency levels.
\item \textbf{System Analysis:} Providing a unified lens for comparing and categorizing existing LLM-HAS implementations.
\item \textbf{Responsible Deployment:} Ensuring that human oversight and control are appropriately calibrated to task requirements, particularly in high-stakes domains.
\end{itemize}

\appsubsection{Definition}
The human agency scale defines five levels (A1–A5) based on the degree of human involvement required for effective task completion (as shown in Table~\ref{tab:haas_levels}).

Different human agency scale levels suit different situations depending on various factors. For example, routine, well-structured tasks may be suitable for A1–A2, while open-ended or creative tasks benefit from A3–A5. Tasks requiring specialized domain knowledge or tacit expertise typically demand higher human involvement (A4–A5), as do high-stakes decisions in healthcare, finance, or legal domains where human oversight and accountability are essential. 

\appsubsection{Human Agency Scale for Design and Analysis}
The Human agency scale framework is not merely a descriptive tool, it also serves as a practical guide for system design. By identifying the appropriate agency level for a given task, designers can make informed decisions about interaction types, feedback mechanisms, and orchestration strategies (Discussed in Section~\ref{sec:components}). Conversely, observing a system's component configuration allows researchers to infer its effective human agency level.

Each human agency level implies distinct requirements for system configuration. At lower levels (A1–A2), systems typically employ delegation-based interaction, asynchronous orchestration, and centralized communication with minimal human touchpoints. Additional, feedback tends to be evaluative and occurs post-task~\cite{yao2022react,xie2024travelplanner,liuagentbench}. As human agency increases toward A3, cooperation and coordination become the dominant interaction patterns, with feedback shifting to guidance and corrective types during task execution. Systems at this level often balance synchronous and asynchronous modes and adopt decentralized communication to facilitate equal partnership~\cite{shao2024collaborative,feng-etal-2024-large,Barres20252BenchEC}. At higher levels (A4–A5), supervision emerges as the primary interaction type, requiring synchronous orchestration and continuous feedback loops. Communication structures become hierarchical with richer interaction modes such as conversation and observation to support sustained human engagement~\cite{qiu2025emoagent}.

\appsection{Empirical Distributions of Core Taxonomy Dimensions}
\label{app:dist}

\begin{figure*}[!t]
    \centering
    \includegraphics[width=\textwidth]{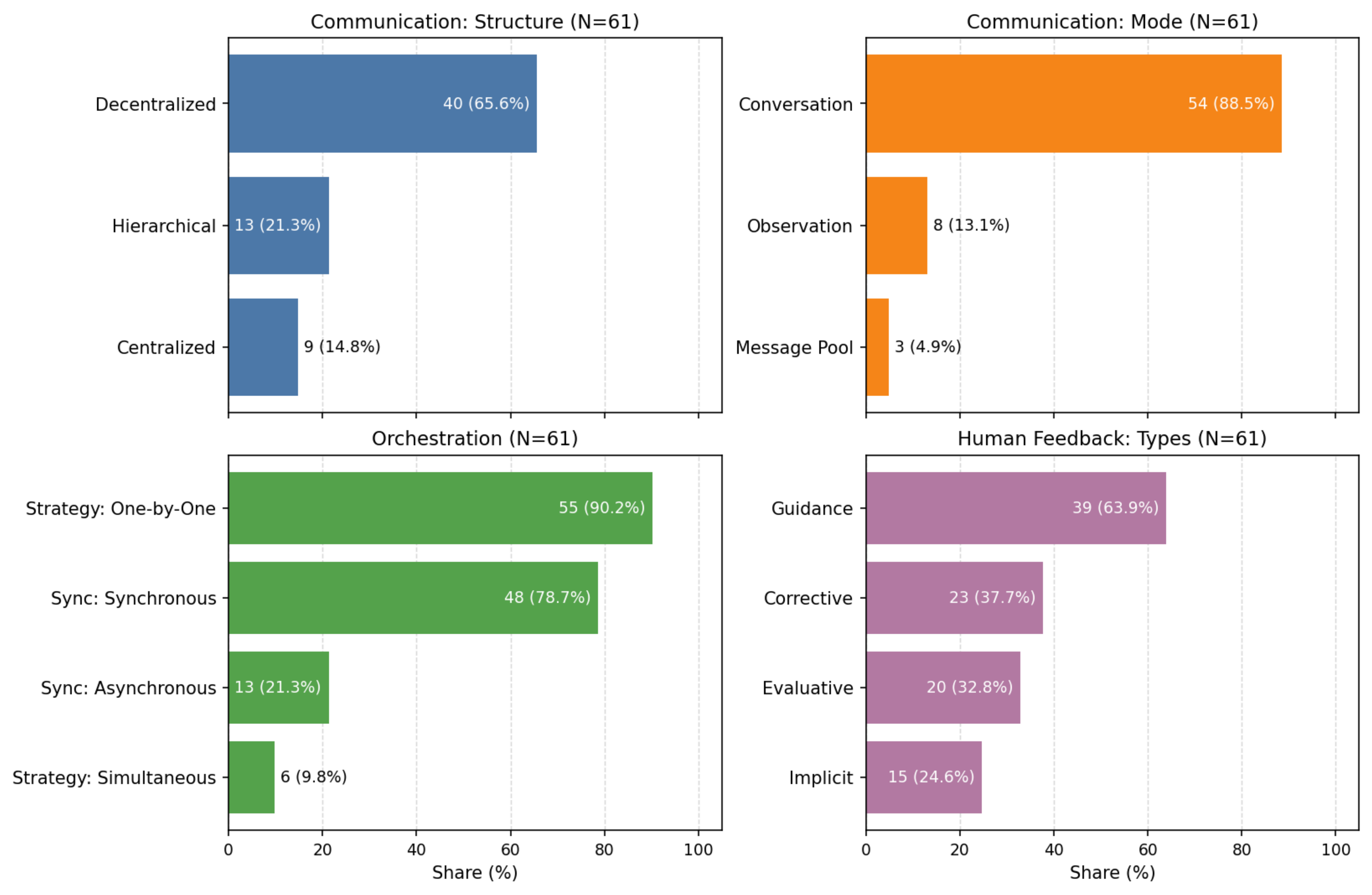}
    \caption{Distributions of core LLM-HAS design dimensions of the collected papers ($N=61$). Bar labels indicate paper count followed by percentage.}
    \label{fig:dist-summary}
\end{figure*}

In this section, we provide quantitative analysis and empirical distribution of communication structures, orchestration paradigms, and human feedback types across the surveyed literature.


\paragraph{Communication Patterns.} The empirical evidence indicates a strong preference for flexible and direct interaction architectures. As shown in Figure~\ref{fig:dist-summary}, Communication Structure is predominantly Decentralized (65.6\%, $N=40$), significantly outpacing Hierarchical (21.3\%) and Centralized (14.8\%) arrangements. This suggests a trend towards autonomous agent behaviors rather than rigid command-and-control topologies from the conventional setting. This decentralization is mirrored in the Communication Mode, where Conversation is the overwhelming standard, utilized in 88.5\% ($N=54$) of the surveyed papers. This indicates that current LLM-HAS designs focus on natural language dialogue, mimicking human interpersonal interactions, over more systemic approaches like Observation (13.1\%) or shared Message Pools (4.9\%).

\paragraph{Orchestration Strategies.} In terms of Orchestration, the landscape is heavily skewed towards linear, sequential workflows. A substantial 90.2\% ($N=55$) of works employ a One-by-One strategy, whereas only 9.8\% attempt Simultaneous execution. This sequential bias is reinforced by the synchronization protocols, with 78.7\% ($N=48$) of systems operating Synchronously. These figures suggest that, despite the potential for parallel processing in LLM agents, current human-agent workflows are designed to align with the linear, turn-taking cognitive limitations of human collaborators, rather than leveraging asynchronous (21.3\%) or concurrent operations. 

\paragraph{Human Feedback Dynamics.} The distribution of Human Feedback highlights that humans are primarily viewed as directors. Guidance is the most prevalent type, appearing in 63.9\% ($N=39$) of works, indicating that humans are frequently involved in proactively steering task execution. In contrast, Corrective (37.7\%) and Evaluative (32.8\%) feedback are less frequent, suggesting that while error recovery and final grading are essential, the core value proposition of humans in current loops is to provide intermediate direction. Implicit feedback remains the least utilized category at 24.6\%, pointing to a significant, under-explored opportunity for systems to learn from passive human signals without demanding explicit user effort.

\appsection{Evaluation Metrics}
In this section, we introduce evaluation metrics specifically designed for human-agent systems across four key aspects: feedback mechanisms, adaptability, trust and safety, and communication methods. To evaluate feedback mechanisms, \cite{liu2024effect} assesses a human-robot teaming framework using multi-modal language feedback at varying frequency levels (inactive, passive, active, superactive). \cite{metz2024mapping} proposes seven metrics, expressiveness, ease, definiteness, context independence, precision, unbiasedness, and informativeness, to evaluate feedback quality. In the education domain, \cite{sessler2025towards} adopts six dimensions based on educational feedback theory. \cite{spencer2020learning} evaluates the Expert Intervention Learning (EIL) method by comparing robot performance with and without expert intervention. For adaptability, \cite{hauptman2023adapt} examines how human-LLM agents respond to cyber incidents under different levels of autonomy across five NIST-defined phases. For trust and safety, \cite{levy2024st} introduces a benchmark that evaluates web agents on their ability to comply with policies, avoid unsafe behavior, respect security constraints, and handle errors gracefully, including seeking user input when needed. Finally, \cite{karten2023interpretable} assess four categories of communication methods in human-agent teaming, focusing on effectiveness and interpretability within simulated environments of Predator-Prey\cite{lowe2017multi} and Traffic Junction\cite{singh2018learning}.

In addition to these aspects, AXIS \cite{lu2024turn} and SYNERGAI \cite{chen2024synergai} evaluate the effectiveness and robustness of human–LLM agent systems in the domains of operating systems and embedded AI, respectively. These studies highlight how evaluation criteria can vary significantly depending on the specific task or application context, reflecting differences in system constraints, performance expectations, and interaction complexity.


\appsection{Ethical and Societal Issues}

Although LLM-based human-agent systems have demonstrated impressive capabilities in different fields, there are still some unresolved social and ethical issues. These problems do not stem purely from model behavior, but from the process of agents interacting with humans and transmitting information. Over time, agents will subtly influence human cognition, emotions, and behavior.

\paragraph{Emotional connection and dependence.} One point of concern is that LLM-based human-agent systems can establish emotional connections with users, allowing people to have emotional projections and trust in agents similar to those between people \citep{cohn2024believing}. As agents are increasingly able to maintain long-term, emotionally charged interactions, users may begin to anthropomorphize them or view them as social partners. Recent empirical studies have shown that while users report an increase in sense of support and engagement when interacting with artificial intelligence partners, such relationships may also weaken real-world social connections and exacerbate loneliness or emotional dependence, especially among socially isolated people \citep{pataranutaporn2025my} . These risks suggest that we need to be wary of users' over-reliance and unrealistic expectations on intelligent agents and balance the boundaries between human and agent interaction and real social interaction.

\paragraph{Responsibility gaps and ambiguous autonomy.} LLM-driven agents often act with partial autonomy, planning and executing tasks without full human oversight and participation. As these systems become more capable, it becomes increasingly difficult to separate the user’s intent from the agent’s autonomous behavior or to assign blame when things go wrong \citep{zou2025call, mukherjee2025stochastic}. This problem means that harm may occur without a clearly identifiable responsible party. If errors or harmful results occur, it is often difficult to clearly identify the responsible party. In most current LLM-HAS architectures, such mechanisms are still inadequate or missing. Solving this problem requires systematic efforts in interpretability, procedural transparency, and governance standards.

\paragraph{Privacy and data-protection risks.} Because LLMs' generative outputs rely on extensive training corpora and user inputs, they have the potential to leak private information. Sensitive information, including identity numbers or medical records, may unintentionally be replicated in generated responses due to the generative nature of these models, according to a recent survey of LLM-based agents. When data moves through several modules, such as the core LLM controller, multi-source inputs, and long-term memory, privacy risks are increased in the long and complex agentic workflow. Sensitive information may be disseminated to other users or outside tools as a result of these components' unregulated data flow. Therefore, strong protections such as strict data usage, safeguards \citep{huang2025deepresearchguard}, and processing filters are necessary to stop LLM agents from disclosing private information. These issues highlight the need for a comprehensive strategy for developing, implementing, and policing LLM-based human-agent systems. It is possible to ensure that such agents promote rather than undermine human well-being by paying close attention to the emotional effects, establishing explicit accountability structures, and enforcing strict privacy protections.

\appsection{Human Feedback Type and Subtype}
\label{sec:human feedback and subtype}

In this appendix, we present a detailed overview of human feedback types and their subtypes, as summarized in Table~\ref{tab:human_feedback_type}. This table provides concise definitions and illustrates how humans provide feedback to LLM-based agents in LLM-HAS. While the main paper introduced the broad categories of evaluative, corrective, guidance, and implicit feedback, here we expand each category into more granular subtypes, ranging from scalar ratings and preference rankings to direct edits, demonstrations, and inferred behavioral signals. Recognizing these subtypes clarifies the ways in which humans interact with LLM agents, by offering precise instructions and well-defined tasks, to enhance the accuracy and quality of generated outputs. This deeper understanding empowers users to optimize their interactions with LLM-based agents. Additionally, the systematic breakdown of human feedback provides a foundation for cross-study comparisons. It underscores the diverse strategies through which human users can guide, correct, or collaborate with LLM-based agents in a more detailed way.

\renewcommand{\arraystretch}{1.4}

\definecolor{headercolor}{RGB}{70, 130, 180}
\definecolor{evaluative}{RGB}{210, 180, 222}
\definecolor{evaluativesub}{RGB}{237, 226, 246}
\definecolor{corrective}{RGB}{174, 214, 241}
\definecolor{correctivesub}{RGB}{214, 234, 248}
\definecolor{guidance}{RGB}{169, 223, 191}
\definecolor{guidancesub}{RGB}{212, 239, 223}
\definecolor{implicit}{RGB}{249, 203, 156}
\definecolor{implicitsub}{RGB}{252, 229, 205}

\begin{table*}[!t]
\centering
\fontsize{8.5pt}{10pt}\selectfont
\begin{tabularx}{\textwidth}{%
>{\raggedright\arraybackslash}p{3.2cm}
>{\raggedright\arraybackslash}p{5.5cm}
>{\raggedright\arraybackslash}p{6cm}}
\toprule
\rowcolor{headercolor}\color{white}\textbf{Human Feedback Type} & \color{white}\textbf{Description} & \color{white}\textbf{How it Helps Agents} \\
\midrule

\rowcolor{evaluative}\textbf{Evaluative Feedback} & User provides an assessment of the agent's output quality. & Signals overall correctness or preference, guiding general alignment. \\
\rowcolor{evaluativesub}\textit{Preference Ranking} & User compares two or more agent outputs and selects the preferred one. & Helps the agent learn relative quality and subjective nuances. \\
\rowcolor{evaluativesub}\textit{Scalar Rating} & User assigns a numerical score (e.g., 1--5) to the agent's output. & Provides a quantitative measure of satisfaction or quality. \\
\rowcolor{evaluativesub}\textit{Binary Assessment} & User indicates simple correctness (e.g., yes/no, thumbs up/down). & Offers a basic signal of success or failure. \\

\rowcolor{corrective}\textbf{Corrective Feedback} & User modifies or directly improves the agent's output. & Provides explicit examples of desired output, enabling direct learning from errors. \\
\rowcolor{correctivesub}\textit{Direct Edits / Refinements} & User manually changes the agent's generated text or code. & Shows the agent the precise correction needed. \\

\rowcolor{guidance}\textbf{Guidance Feedback} & User provides instructions or explanations to steer the agent. & Offers deeper context, reasoning, or demonstrations for learning complex behaviors. \\
\rowcolor{guidancesub}\textit{Demonstrations} & User shows the agent how to perform a task correctly. & Teaches specific procedures or desired interaction patterns. \\
\rowcolor{guidancesub}\textit{Instructions / Critiques} & User provides natural language explanations, critiques, or step-by-step guidance. & Helps the agent understand why an output is wrong and how to improve. \\

\rowcolor{implicit}\textbf{Implicit Feedback} & Agent infers user preference from their behavior. & Reveals preferences and usability issues without explicit feedback requests. \\
\rowcolor{implicitsub}\textit{Human Action / Control} & Human directly takes actions and control. & Collaborates with humans to effectively finish tasks or learns from human actions. \\

\bottomrule
\end{tabularx}
\caption{Human Feedback Types and Subtypes. The subtypes of evaluative feedback includes preference ranking, scalar rating, and binary assessment. The subtypes of corrective feedback includes the direct edits or refinement. The subtypes of guidance feedback includes the demonstration and instructions or critiques. The subtypes of implicit feedback include the human action or control.}
\label{tab:human_feedback_type}
\end{table*}

\appsection{Difference with Traditional Human-in-the-Loop and Human-Computer Interaction Systems}
\label{sec:appendix_difference}

LLM-based human-agent systems (LLM-HAS) differ from traditional human-in-the-loop (HITL) systems and classic human-computer interaction (HCI) frameworks. They vary in system structure, interaction dynamics, and the way they use feedback. Although all three involve human participation, they have different ideas about the role of humans, the independence of intelligent systems, and how collaboration works \citep{wu2022survey, borghoff2025human}.

\paragraph{LLM-HAS vs. HITL.}
Traditional HITL systems often include humans at fixed and predictable stages of the machine learning pipeline, like data labeling, model selection, or post-correction \citep{kim2025because}. Human involvement is usually occasional and specific to tasks, and feedback is mostly gathered offline in structured formats, such as labels, binary corrections, or rankings. Because of this, HITL frameworks focus on control, supervision, and reducing errors but provide little support for ongoing, interactive, or two-way collaboration during task execution. In contrast, LLM-HAS allow continuous, multi-round interaction using natural language. This lets humans guide, critique, refine, or redirect agent behavior as tasks progress. Instead of mainly being supervisors or annotators, humans in LLM-HAS become active collaborators whose input influences both the process and the results of agent actions.

\paragraph{LLM-HAS vs. Traditional HCI Systems.}
Classic HCI systems are often set up for direct manipulation or command-response interaction, where users specify actions that systems respond to in a fixed way. Even though modern HCI research increasingly focuses on user-centered design and interactive experiences, most HCI systems do not see computational components as independent agents with their own initiative or reasoning abilities \citep{xu2023transitioning}. In contrast, LLM-HAS introduce agentic elements that can create plans, start actions, ask for clarification, and change their behavior based on ongoing interactions. This change moves the interaction from simple tool use to collaboration, allowing for smoother and more flexible human-agent workflows that go beyond traditional interface-based interaction models.

\paragraph{Feedback and Adaptation.}
Another major difference is how feedback is represented and used. HITL systems typically depend on infrequent, structured feedback gathered for training or evaluation, while traditional HCI systems often see user feedback as temporary signals that don’t directly affect system behavior during a session. LLM-HAS, on the other hand, can take in rich, natural language feedback in real time. This allows users to express complex intentions, preferences, and judgments as they work on tasks. Supported by large language models, agents in LLM-HAS can learn from minimal input, modify their responses immediately, and change their behavior without needing retraining. This ability for quick adaptation and personalization sets LLM-HAS apart from both HITL and traditional HCI models.

Together, these differences position LLM-HAS as a new type of interactive intelligent system that blends adaptive intelligence with human-centered design. Rather than just enhancing existing HITL or HCI frameworks, LLM-HAS function under a different model where humans and agents work together in reasoning, decision-making, and action throughout the interaction process.

\appsection{Difference with Multi-Agent Systems}
While both LLM-HAS and MAS involve collaboration among multiple entities, the key distinction lies in the nature and role of the collaborating parties \cite{feng-etal-2024-large, shao2024collaborative}. Multi-agent systems are typically composed exclusively of autonomous agents—each designed to make decisions, communicate, and coordinate tasks with one another. In these MAS, each agent operates based on its own set of objectives and algorithms, and the overall behavior emerges from their interactions \cite{tran2025multi, guo2024large}. 

In contrast, LLM-based human–agent systems explicitly incorporate humans as active participants within the decision-making loop \cite{feng-etal-2024-large}. Rather than letting the system run purely on the combined strategies of several LLM-powered agents, these systems are engineered with mechanisms to allow human supervision, intervention, and feedback \cite{mehta-etal-2024-improving}. This human-in-the-loop design is critical when balancing the strengths of LLMs,such as processing vast amounts of knowledge and performing rapid reasoning, with the need for contextual, ethical, and domain-specific judgments that humans uniquely provide \cite{vats2024survey}.

Furthermore, multi-agent systems often assume that the collaboration among agents can lead to a form of “collective intelligence” where agents work toward shared objectives \cite{suninteractgen}. In many such frameworks, the communication protocols, coordination strategies, and role dynamics are all defined among non-human entities. In contrast, in human–agent systems, the interaction protocols are designed to enhance transparency and provide control for human decision-makers \cite{shao2024collaborative}. The system can selectively escalate issues for human review, enable corrective actions when the automated decision may be off-mark, and integrate human feedback to iteratively improve the agent’s performance over time \cite{mehta-etal-2024-improving}.

\appsection{Tables}
 Table~\ref{tab:humanfeedback} catalogs the environmental configuration and human feedback type, and Table~\ref{tab:humancommunication} categorizes the interaction, orchestration, and communication of representative works, respectively. Both tables present the representative work. For all the collected work, please refer to the Github page.
\newcolumntype{C}[1]{>{\centering\arraybackslash}p{#1}}
\newcolumntype{I}[1]{>{\itshape\centering\arraybackslash}p{#1}}
\newcolumntype{L}[1]{>{\raggedright\arraybackslash}p{#1}}

\definecolor{orchestClr}{RGB}{240,216,190}  
\definecolor{interactClr}{RGB}{190,215,255}

\definecolor{communClr}{RGB}{230,210,255}
\colorlet   {interactLight}{interactClr!55}
\colorlet   {orchestLight}{orchestClr!55}
\colorlet   {communLight}{communClr!55}

\definecolor{rowA}{RGB}{248,248,248}
\rowcolors{3}{rowA}{white}

\setlength{\tabcolsep}{0.1pt}
\renewcommand{\arraystretch}{1.35}


\newcommand{\TabScale}{0.83}
\newcommand{\TabScaleNum}{83}
\newcommand{\TabScaleDen}{100}
\newlength{\TabW}
\setlength{\TabW}{\dimexpr\textwidth*\TabScaleDen/\TabScaleNum\relax}

\begin{table*}[t]
\centering
\captionsetup{width=\textwidth}
\captionsetup[table]{justification=raggedright, singlelinecheck=false}
\caption{\ding{172} Environment Configuration and \ding{173} Human Feedback to LLM-based agents in human–agent systems.
Environment Configuration specifies whether a single or multiple humans collaborate with one or more LLM-based agents, while Human Feedback characterizes the type, subtype, granularity, and interaction phase of the human feedback to the LLM-based agents.}
\label{tab:humanfeedback}

{\fontsize{6.5pt}{7.2pt}\selectfont
\setlength{\tabcolsep}{2pt}
\renewcommand{\arraystretch}{1.18}

\scalebox{\TabScale}{%
\begin{tabularx}{\TabW}{@{}%
  L{3.5cm}  I{1.2cm} C{1.4cm}  C{1.5cm} C{1.5cm}
  C{2.3cm} C{3.1cm} C{1.5cm} C{1.9cm}@{}}
\toprule
\multicolumn{3}{c}{} &
\multicolumn{2}{c}{\cellcolor{orchestClr}\textbf{Environment Configuration}} &
\multicolumn{4}{c}{\cellcolor{communClr}\textbf{Human Feedback to LLM-based Agent}}\\
\cmidrule(lr){4-5}\cmidrule(lr){6-9}
\textbf{Paper} & \textbf{Venue} & \textbf{Code/\ Data} &
\cellcolor{orchestLight}\textbf{Human} &
\cellcolor{orchestLight}\textbf{LLM Agent} &
\cellcolor{communLight}\textbf{Type} &
\cellcolor{communLight}\textbf{Subtype} &
\cellcolor{communLight}\textbf{Granularity} &
\cellcolor{communLight}\textbf{Phase}\\
\midrule
\paper{Collaborative Gym}{shao2024collaborative} & \textit{Arxiv'24} & \codelink{https://github.com/SALT-NLP/collaborative-gym} & Single & Single & Corrective, Guidance & Refinements, Instructions & Segment & During Task \\
\paper{MTOM}{Zhang2024MutualTO}  & \textit{Arxiv'24} & \textit{--} & Single & Single & Implicit & Human Action & Segment & During Task \\
\paper{FineArena}{xu2025finarena} & \textit{Arxiv'25} & \textit{--} & Single & Multiple & Guidance & Demonstrations & Segment, Holistic & Initial Setup, During Task \\
\paper{Prison Dilemm}{jiang2025experimentalexplorationinvestigatingcooperative} & \textit{Arxiv'25} & \textit{--} & Single & Single & Implicit & Human Action & Segment & During Task \\
\paper{PPP}{sun2025training} & \textit{Arxiv'25} & \codelink{https://github.com/sunnweiwei/PPP-Agent} & Single & Single & Guidance, Evaluate & Scalar rating, Refinements & Segment, Holistic & During task \\
\paper{AI Chains}{10.1145/3491102.3517582} & \textit{CHI'24} & \textit{--} & Single & Single & Corrective & Refinements & Segment & During Task \\
\paper{Drive As You Speak}{cui2024drive} & \textit{WACV'24} & \textit{--} & Single & Single & Guidance & Demonstrations & Holistic & Initial Setup \\
\paper{AgentCoord}{pan2024agentcoord} & \textit{Arxiv'24} & \codelink{https://github.com/AgentCoord/AgentCoord} & Single & Multiple & Evaluative, Corrective & Preference Ranking, Refinements & Segment, Holistic & Initial Setup, During Task\\
\paper{CowPilot}{huq2025cowpilotframeworkautonomoushumanagent} & \textit{Arxiv'25} & \codelink{https://oaishi.github.io/cowpilot.html} & Single & Single & Corrective, Evaluative & Binary Assessment, Refinements & Segment & During Task \\
\paper{EasyLAN}{pan2024human} & \textit{Arxiv'24} & \textit{--} & Single & Multiple & Corrective, Guidance & Demonstrations, Refinements & Segment, Holistic & During Task \\
\paper{Hierarchical Agent}{Liu2023LLMPoweredHL} & \textit{AAMAS'24} & \textit{--} & Single & Multiple & Guidance & Demonstrations & Segment & During Task \\
\paper{SWEET‑RL}{zhou2025sweetrltrainingmultiturnllm} & \textit{Arxiv'25} & \codelink{https://github.com/facebookresearch/sweet_rl} & Single & Single & Corrective, Implicit & Refinements, Human Action & Segment & Initial Setup, During Task \\
\paper{HRC Assembly}{gkournelos2024llm} & \textit{CIRP'24} & \textit{--} & Single & Multiple & Guidance & Demonstrations & Segment & During Task \\
\paper{REVECA}{seo2025reveca} & \textit{Arxiv'24} & \textit{--} & Single & Multiple & Guidance & Demonstrations & Holistic & Initial Setup \\
\paper{AssistantX}{sun2024assistantx} & \textit{Arxiv'24} & \codelink{https://github.com/AssistantX-Agent/AssistantX} & Multiple & Multiple & Implicit, Guidance & Human Action, Demonstrations & Holistic, Segment & Initial Setup, During Task\\
\paper{MINT}{wang2024mint} & \textit{ICLR'24} & \codelink{https://xwang.dev/mint-bench/} & Multiple & Single & Evaluative, Corrective, Guidance & Binary Assessment, Refinements, Instructions & Holistic & During Task \\
\paper{Help Feedback}{mehta-etal-2024-improving} & \textit{EACL'24} & \textit{--} & Single & Single & Evaluative, Guidance & Demonstrations, Instructions, Binary Assessment & Holistic, Segment & During Task \\
\paper{ConvCodeWorld}{han2025convcodeworld} & \textit{ICLR'25} & \codelink{https://github.com/stovecat/convcodeworld} & Single & Single & Guidance, Evaluative & Demonstrations, Instructions, Binary Assessment & Segment, Holistic & During Task \\
\paper{ReHAC}{feng-etal-2024-large} & \textit{ACL'24} & \codelink{https://github.com/XueyangFeng/ReHAC} & Single & Single & Corrective & Refinements & Segment & During Task \\
\paper{DPT Agent}{zhang2025leveraging} & \textit{Arxiv'25} & \codelink{https://github.com/sjtu-marl/DPT-Agent} & Single & Single & Guidance & Instructions & Holistic & During Task \\
\paper{HRC Manipulation}{liu2023llm} & \textit{IEEE'23} & \textit{--} & Single & Single & Corrective, Guidance & Demonstrations, Refinements & Segment & During Task \\
\paper{HRC DMP}{liu2024enhancing} & \textit{IEEE'24} & \textit{--} & Single & Single & Corrective, Guidance & Refinements, Demonstrations & Segment & During Task \\
\paper{PARTNR}{chang2024partnr} & \textit{ICLR'25} & \codelink{https://github.com/facebookresearch/partnr-planner/tree/main/} & Single & Single & Guidance & Demonstrations & Holistic & Inital Setup\\
\paper{Organized Teams}{guo2024embodiedllmagentslearn} & \textit{Arxiv'24} & \codelink{https://github.com/tobeatraceur/Organized-LLM-Agents} & Single & Multiple & Guidance & Demonstrations & Holistic, Segment & Initial Setup, During Task \\
\paper{CoELA}{zhang2024building} & \textit{ICLR'23} & \textit{--} & Single & Multiple & Guidance & Demonstrations & Holistic, Segment & Initial Setup, During Task \\
\paper{Agency Task}{sharma2024investigatingagencyllmshumanai} & \textit{EACL'24} & \codelink{https://github.com/microsoft/agency-dialogue} & Single & Single & Guidance & Demonstrations & Segment & During Task \\
\paper{GDfC}{wang2025human} & \textit{SME'25} & \textit{--} & Single & Multiple & Guidance, Evaluative & Demonstrations, Binary Assessment, Preference Ranking & Holistic, Segment & Initial Setup, During Task, Post Task\\
\paper{PDFChatAnnotator}{10.1145/3640543.3645174} & \textit{IUI'24} & \textit{--} & Single & Single & Corrective, Guidance & Demonstrations, Refinements & Segment & During Task \\
\paper{Attentive Supp.}{tanneberg2024help} & \textit{IEEE'24} & \codelink{https://github.com/HRI-EU/AttentiveSupport} & Multiple & Single & Implicit, Guidance & Demonstrations, Human Action & Segment & During Task\\
\paper{HRC Trust}{10141597} & \textit{IEEE'23} & \textit{--} & Single & Single & Guidance & Demonstrations, Instructions & Segment & During Task \\
\paper{BPMN}{ait2024towards} & \textit{Arxiv'24} & \codelink{https://github.com/BESSER-PEARL/agentic-bpmn} & Multiple & Multiple & Guidance, Corrective & Instructions, Refinements & Segment &  During Task, Post Task\\
\paper{Co‑STORM}{jiang-etal-2024-unknown} & \textit{EMNLP'24} & \codelink{https://github.com/stanford-oval/storm} & Single & Multiple & Guidance & Demonstrations & Segment & During Task \\
\paper{HRC Manufa.}{10711843} & \textit{IEEE'24} & \textit{--} & Single & Single & Corrective, Guidance & Demonstrations, Refinements, Instructions & Segment & Initial Setup, During Task\\
\paper{A2C}{tariq2025a2c} & \textit{Arxiv'24} & \codelink{https://anonymous.4open.science/r/A2C/README.md} & Multiple & Multiple & Guidance, Evaluative & Binary Assessment, Instructions & Holistic, Segment & During Task \\
\paper{MindAgent}{gong2023mindagent} & \textit{NAACL'24} & \codelink{https://github.com/mindagent/mindagent} & Single & Multiple & Guidance & Demonstrations & Segment & During Task \\
\paper{Ask Before Plan}{zhang-etal-2024-ask} & \textit{EMNLP'24} & \codelink{https://github.com/magicgh/Ask-before-Plan} & Single & Multiple & Guidance & Demonstrations & Segment & Initial Setup, During Task \\

\paper{SOTOPIA}{zhou2024sotopia} & \textit{ICLR'24} & \textit{--} & Multiple & Multiple & Evaluative, Implicit & Scaler Rating, Human Action & Holistic, Segment & During Task, Post Task\\

\paper{PaLM‑E}{10.5555/3618408.3618748} & \textit{ICML'23} & \codelink{https://palm-e.github.io} & Single & Single & Guidance, Implicit & Demonstrations, Human Action & Holistic, Segment & Initial Setup, During Task \\
\paper{TaPA}{wu2023embodied} & \textit{Arxiv'23} & \codelink{https://github.com/Gary3410/TaPA} & Single & Single & Guidance & Demonstrations & Holistic, Segment & Initial Setup \\
\paper{MetaGPT}{hong2023metagpt} & \textit{ICLR'24} & \codelink{https://github.com/geekan/MetaGPT} & Single & Multiple & Guidance & Demonstrations & Holistic & Initial Setup\\
\paper{DigiRL}{NEURIPS2024_1704ddd0} & \textit{NeurIPS'24} & \codelink{https://github.com/DigiRL-agent/digirl} & Single & Single & Evaluative, Guidance & Binary Assessment, Demonstrations & Holistic & During Task, Post Task\\
\paper{WebLINX}{lu2024weblinx} & \textit{Arxiv'24} & \codelink{https://github.com/McGill-NLP/WebLINX} & Single & Multiple & Guidance & Demonstrations & Holistic, Segment & Initial Setup, During Task \\

\paper{MineWorld}{guo2025mineworld} & \textit{Arxiv'25} & \codelink{https://github.com/microsoft/MineWorld} & Multiple & Single & Implicit & Human Action & Segment & During Task \\

\paper{M3HF}{wang2025m3hf} & \textit{ICML'25} & \textit{--} & Multiple & Multiple & Evaluative, Guidance & Binary Assessment, Instructions & Segment, Holistic & During Task, Post Task \\

\paper{UserBench}{Qian2025UserBenchAI} & \textit{Arxiv'25} & \codelink{https://github.com/SalesforceAIResearch/UserBench} & Single & Single & Implicit, Guidance & Human Action, Refinement & Segment & Initial Setup, During Task \\

\paper{$\tau^2$-Bench}{Barres20252BenchEC} & \textit{Arxiv'25} & \codelink{https://github.com/sierra-research/tau2-bench} & Single & Single & Evaluative, Implicit & Human Action, Binary assessment & Segment, Holistic & Initial Setup, During Task \\

\paper{Magentic-UI}{Mozannar2025MagenticUITH} & \textit{Arxiv'24} & \codelink{https://github.com/microsoft/magentic-ui} & Single & Multiple & Evaluative, Corrective, Guidance, Implicit & Binary Assessment, Refinement, Corrective, Human Action & Segment & During Task, Post Task \\

\paper{RECODE-H}{miao2025recode} & \textit{Arxiv'25} & \codelink{https://github.com/ChunyuMiao98/RECODE-H} & Single & Single &  Guidance, Corrective & Refinements, Corrective, Demonstration & Segment & During Task \\

\paper{EmoAgent}{qiu2025emoagent} & \textit{Arxiv'25} & \textit{--} & Single & Multiple &  Corrective, Implicit, Guidance & Human Action, Instructions, Binary Assessment & Segment, Holistic & During Task, Post Task \\

\paper{SymbioticRAG}{sun2025symbioticrag} & \textit{Arxiv'25} & \textit{--} & Single & Single &  Corrective, Implicit, Evaluative & Binary Assessment, Refinements, Demonstrations, Instructions, Human Action & Segment & Initial Setup, During Task, Post Task \\

\bottomrule
\end{tabularx}%
} 
} 
\end{table*}
\newcolumntype{C}[1]{>{\centering\arraybackslash}p{#1}}
\newcolumntype{I}[1]{>{\itshape\centering\arraybackslash}p{#1}}
\newcolumntype{L}[1]{>{\raggedright\arraybackslash}p{#1}}

\definecolor{interactClr}{RGB}{190,215,255}
\definecolor{communClr}{RGB}{206,240,220}
\colorlet   {orchestLight}{orchestClr!55}

\definecolor{orchestClr}{RGB}{255,196,141}  

\colorlet   {interactLight}{interactClr!55}
\colorlet   {orchestLight}{orchestClr!55}
\colorlet   {communLight}{communClr!55}

\definecolor{rowA}{RGB}{248,248,248}
\rowcolors{3}{rowA}{white}

\setlength{\tabcolsep}{0.1pt}
\renewcommand{\arraystretch}{1.6}

\setlength{\TabW}{\dimexpr\textwidth*\TabScaleDen/\TabScaleNum\relax}

\begin{table*}[t]
\centering
\captionsetup{width=\textwidth}
\captionsetup[table]{justification=raggedright,singlelinecheck=false}
  \caption{%
    \ding{172} Interaction  
    \ding{173} Orchestration  
    \ding{174} Communication in LLM-based human–agent systems.  
    Interaction types capture the human and agent collaboration type; Orchestration covers task strategy and temporal synchronization; Communication describes how messages are structured and delivered in the system.
  }
\label{tab:humancommunication}
{\fontsize{6.5pt}{7.2pt}\selectfont
\setlength{\tabcolsep}{2pt}
\renewcommand{\arraystretch}{1.18}
\scalebox{\TabScale}{%
\begin{tabularx}{\TabW}{@{}%
  L{3.5cm}  I{1.2cm} C{1.4cm}  C{1.5cm} C{1.5cm}
  C{2.3cm} C{3.1cm} C{1.5cm} C{1.9cm}@{}}
\toprule
\multicolumn{3}{c}{} &
\multicolumn{2}{c}{\cellcolor{interactClr}\textbf{Interaction}} &
\multicolumn{2}{c}{\cellcolor{orchestClr}\textbf{Orchestration}} &
\multicolumn{2}{c}{\cellcolor{communClr}\textbf{Communication}}\\
\cmidrule(lr){4-5}\cmidrule(lr){6-7}\cmidrule(lr){8-9}
\textbf{Paper} & \textbf{Venue} & \textbf{Code/\ Data} &
\cellcolor{interactLight}\textbf{Types} & \cellcolor{interactLight}\textbf{Variant} &
\cellcolor{orchestLight}\textbf{Strategy} & \cellcolor{orchestLight}\textbf{Synchronization} &
\cellcolor{communLight}\textbf{Structure} & \cellcolor{communLight}\textbf{Mode}\\
\midrule
\paper{Collaborative Gym}{shao2024collaborative} & Arxiv'24 & \codelink{https://github.com/SALT-NLP/collaborative-gym} & Collaboration & Cooperation, Delegation & One‑by‑One & Asynchronous & Decentralized & Conversation\\
\paper{MTOM}{Zhang2024MutualTO}  & Arxiv'24 & \textit{--}                               & Collaboration & Coordination, Cooperation                               & Simultaneous& Synchronous  & Decentralized & Conversation\\
\paper{FineArena}{xu2025finarena} & Arxiv'25 & \textit{--} & Collaboration & Delegation, Cooperation & One‑by‑One & Synchronous  & Hierarchical  & Conversation\\
\paper{Prison Dilemm}{jiang2025experimentalexplorationinvestigatingcooperative} & Arxiv'25 & \textit{--} & Coopetition   & -- & One‑by‑One & Asynchronous  & Decentralized & Conversation\\
\paper{PPP}{sun2025training} & \textit{Arxiv'25} & \codelink{https://github.com/sunnweiwei/PPP-Agent} & Collaboration & Cooperation       & One‑by‑One & Asynchronous & Decentralized & Conversation\\
\paper{AI Chains}{10.1145/3491102.3517582} & CHI'24   & \textit{--} & Collaboration & Cooperation & One‑by‑One & Synchronous  & Decentralized  & Conversation\\
\paper{Drive As You Speak}{cui2024drive}  & WACV'24  & \textit{--} & Collaboration & Delegation & One‑by‑One & Synchronous  & Centralized   & Conversation\\
\paper{AgentCoord}{pan2024agentcoord}  & Arxiv'24 & \codelink{https://github.com/AgentCoord/AgentCoord} & Collaboration & Coordination  & One‑by‑One & Synchronous  & Hierarchical & Conversation\\
\paper{CowPilot}{huq2025cowpilotframeworkautonomoushumanagent} & Arxiv'25 & \codelink{https://oaishi.github.io/cowpilot.html} & Collaboration & Supervision, Delegation, Coordination & One‑by‑One & Synchronous & Decentralized & Conversation\\
\paper{EasyLAN}{pan2024human}  & Arxiv'24 & \textit{--} & Collaboration & Delegation, Supervision & One‑by‑One & Synchronous  & Hierarchical  & Observation\\
\paper{Hierarchical Agent}{Liu2023LLMPoweredHL}& AAMAS'24 & \textit{--}   & Collaboration & Supervision, Delegation, Cooperation & One-by-One & Synchronous & Hierarchical & Conversation\\
\paper{SWEET‑RL}{zhou2025sweetrltrainingmultiturnllm} & Arxiv'25 & \codelink{https://github.com/facebookresearch/sweet_rl} & Collaboration & Delegation & One‑by‑One & Synchronous  & Centralized & Conversation\\
\paper{HRC Assembly}{gkournelos2024llm}& CIRP'24  & \textit{--}  & Collaboration & Delegation, Cooperation & One‑by‑One & Synchronous  & Decentralized   & Conversation\\
\paper{REVECA}{seo2025reveca} & Arxiv'24 & \textit{--} & Collaboration & Cooperation & One‑by‑One & Synchronous & Decentralized  & Conversation\\
\paper{AssistantX}{sun2024assistantx}  & Arxiv'24 & \codelink{https://github.com/AssistantX-Agent/AssistantX} & Collaboration & Delegation, Cooperation  & One‑by‑One & Asynchronous & Decentralized & Message Pool\\
\paper{MINT}{wang2024mint} & ICLR'24  & \codelink{https://xwang.dev/mint-bench/} & Collaboration & Delegation, Cooperation & One‑by‑One & Synchronous  & Decentralized & Conversation\\
\paper{Help Feedback}{mehta-etal-2024-improving} & EACL'24  & \textit{--} & Collaboration & Supervision, Delegation, Cooperation & One‑by‑One & Asynchronous  & Decentralized & Conversation\\
\paper{ConvCodeWorld}{han2025convcodeworld} & ICLR'25  & \codelink{https://github.com/stovecat/convcodeworld} & Collaboration & Supervision, Delegation & One‑by‑One & Asynchronous & Decentralized & Conversation\\
\paper{ReHAC}{feng-etal-2024-large}  & ACL'24   & \codelink{https://github.com/XueyangFeng/ReHAC} & Collaboration & Coordination, Supervision & One‑by‑One & Synchronous  & Decentralized & Conversation\\
\paper{DPT Agent}{zhang2025leveraging} & Arxiv'25 & \codelink{https://github.com/sjtu-marl/DPT-Agent} & Collaboration & Coordination & Simultaneous& Asynchronous & Decentralized & Observation\\
\paper{HRC Manipulation}{liu2023llm}  & IEEE'23  & \textit{--} & Collaboration & Supervision, Delegation  & One‑by‑One & Synchronous  & Decentralized & Conversation\\
\paper{HRC DMP}{liu2024enhancing} & IEEE'24  & \textit{--} & Collaboration & Delegation, Supervision & One‑by‑One & Synchronous  & Decentralized & Conversation\\
\paper{PARTNR}{chang2024partnr} & ICLR'25  & \codelink{https://github.com/facebookresearch/partnr-planner/tree/main/} & Collaboration & Coordination, Cooperation & Simultaneous & Synchronous & Decentralized, Centralized  & Observation\\
\paper{Organized Teams}{guo2024embodiedllmagentslearn} & Arxiv'24 & \codelink{https://github.com/tobeatraceur/Organized-LLM-Agents} & Collaboration & Cooperation, Coordination & One‑by‑One & Synchronous  & Decentralized, Centralized, Hierarchical & Conversation\\
\paper{CoELA}{zhang2024building} & ICLR'23  & \textit{--} & Collaboration & Cooperation, Coordination & Simultaneous& Synchronous & Decentralized & Conversation\\
\paper{Agency Task}{sharma2024investigatingagencyllmshumanai} & EACL'24  & \codelink{https://github.com/microsoft/agency-dialogue}  & Collaboration & Cooperation, Delegation & One‑by‑One & Synchronous  & Decentralized & Conversation\\
\paper{GDfC}{wang2025human} & SME'25 & \textit{--} & Collaboration & Delegation & One‑by‑One & Synchronous  & Decentralized & Conversation\\
\paper{PDFChatAnnotator}{10.1145/3640543.3645174} & IUI'24  & \textit{--} & Collaboration & Delegation & One‑by‑One & Synchronous  & Decentralized & Conversation\\
\paper{Attentive Supp.}{tanneberg2024help} & IEEE'24  & \codelink{https://github.com/HRI-EU/AttentiveSupport} & Collaboration & Coordination & One‑by‑One & Synchronous  & Decentralized & Observation\\

\paper{HRC Trust}{10141597} & IEEE'23  & \textit{--} & Collaboration & Delegation & One‑by‑One & Synchronous  & Decentralized & Conversation\\

\paper{BPMN}{ait2024towards}  & Arxiv'24 & \codelink{https://github.com/BESSER-PEARL/agentic-bpmn} & Collaboration & Coordination  & Simultaneous & Asynchronous & Decentralized & Message Pool\\

\paper{Co-STORM}{jiang-etal-2024-unknown} & EMNLP'24 & \codelink{https://github.com/stanford-oval/storm} & Collaboration & Coordination & One‑by‑One & Synchronous & Centralized & Conversation\\

\paper{HRC Manufa.}{10711843}  & IEEE'24  & \textit{--} & Collaboration & Delegation, Cooperation & One‑by‑One & Synchronous  & Centralized & Conversation\\

\paper{A2C}{tariq2025a2c}  & Arxiv'24 & \codelink{https://anonymous.4open.science/r/A2C/README.md} & Collaboration & Cooperation & One‑by‑One & Asynchronous & Hierarchical & Conversation\\

\paper{MindAgent}{gong2023mindagent} & NAACL'24 & \codelink{https://github.com/mindagent/mindagent} & Collaboration & Coordination & Simultaneous & Synchronous & Centralized & Conversation\\

\paper{Ask Before Plan}{zhang-etal-2024-ask} & EMNLP'24 & \codelink{https://github.com/magicgh/Ask-before-Plan} & Collaboration & Coordination, Delegation & One‑by‑One & Synchronous & Hierarchical & Conversation\\

\paper{SOTOPIA}{zhou2024sotopia} & ICLR'24  & \textit{--}                               & Collaboration, Competition, Coopetition & Coordination, Cooperation   & One‑by‑One & Synchronous & Decentralized & Conversation\\

\paper{PaLM‑E}{10.5555/3618408.3618748}  & ICML'23  & \codelink{https://palm-e.github.io} & Collaboration & Delegation   & One‑by‑One & Synchronous  & Decentralized & Conversation\\

\paper{TaPA}{wu2023embodied}  & Arxiv'23 & \codelink{https://github.com/Gary3410/TaPA} & Collaboration & Delegation   & One‑by‑One & Asynchronous & Decentralized & Conversation\\

\paper{MetaGPT}{hong2023metagpt}  & ICLR'24  & \codelink{https://github.com/geekan/MetaGPT} & Collaboration & Coordination & One‑by‑One & Asynchronous & Decentralized & Message Pool\\

\paper{DigiRL}{NEURIPS2024_1704ddd0}  & NeurIPS'24&\codelink{https://github.com/DigiRL-agent/digirl} & Collaboration & Delegation & One‑by‑One & Synchronous & Centralized & Conversation\\

\paper{WebLINX}{lu2024weblinx} & Arxiv'24 & \codelink{https://github.com/McGill-NLP/WebLINX} & Collaboration & Delegation & One‑by‑One & Synchronous  & Hierarchical & Conversation\\

\paper{MineWorld}{guo2025mineworld} & Arxiv'25 & \codelink{https://github.com/microsoft/MineWorld} & Collaboration & Delegation & One‑by‑One & Synchronous  & Decentralized & Observation\\

\paper{M3HF}{wang2025m3hf} & \textit{ICML'25} & \textit{--} & Collaboration & Cooperation & One-by-One, Simultaneous & Synchronous & Centralized & Message Pool \\

\paper{UserBench}{Qian2025UserBenchAI} & \textit{Arxiv'25} & \codelink{https://github.com/SalesforceAIResearch/UserBench} & Collaboration & Cooperation & One-by-One & Asynchronous & Decentralized & Conversation \\ 

\paper{$\tau^2$-Bench}{Barres20252BenchEC} & \textit{Arxiv'25} & \codelink{https://github.com/sierra-research/tau2-bench} & Collaboration & Cooperation, Coordination & One-by-One, Simultaneous & Synchronous & Decentralized, Hierarchical & Conversation \\

\paper{Magentic-UI}{Mozannar2025MagenticUITH} & \textit{Arxiv'24} & \codelink{https://github.com/microsoft/magentic-ui} & Collaboration & Cooperation, Coordination & Simultaneous & Asynchronous, Synchronous & Hierarchical, Centralized & Conversation, Observation \\

\paper{RECODE-H}{miao2025recode} & \textit{Arxiv'25} & \codelink{https://github.com/ChunyuMiao98/RECODE-H} & Collaboration & Supervision, Cooperation & One-by-One & Synchronous & Hierarchical & Conversation \\

\paper{EmoAgent}{qiu2025emoagent} & \textit{Arxiv'25} & \textit{--} & Collaboration & Supervision, Coordination, Cooperation & One-by-One & Synchronous & Hierarchical, Centralized & Conversation, Observation \\

\paper{SymbioticRAG}{sun2025symbioticrag} & \textit{Arxiv'25} & \textit{--} & Collaboration & Cooperation, Supervision, Delegation & One-by-One & Synchronous & Centralized & Conversation \\

\bottomrule
\end{tabularx}%
} 
} 
\end{table*}





\end{document}